\documentclass[10pt,twocolumn,letterpaper]{article}
\pdfoutput=1 
\usepackage{iccv}
\usepackage{times}
\usepackage{epsfig}
\usepackage{graphicx}
\usepackage{amsmath}
\usepackage{amssymb}
\usepackage{subfigure}
\usepackage{booktabs}
\usepackage{multirow}
\usepackage[ruled]{algorithm2e}

\usepackage[pagebackref=true,breaklinks=true,letterpaper=true,colorlinks,bookmarks=false]{hyperref}

\iccvfinalcopy % *** Uncomment this line for the final submission

\def\iccvPaperID{3100} % *** Enter the ICCV Paper ID here

\ificcvfinal\fi

\begin{document}

%%%%%%%%% TITLE
\title{Greedy Gradient Ensemble for Robust Visual Question Answering}

\author{Xinzhe Han$^{2,1}$ ~~ Shuhui Wang$^{1}$\thanks{Corresponding author.} ~~ Chi Su$^{3}$ ~~ Qingming Huang$^{1,2,4}$ ~~ Qi Tian$^{5}$\\
	$^{1}$Key Lab of Intell. Info. Process., Inst. of Comput. Tech., CAS, Beijing, China\\
	$^{2}$University of Chinese Academy of Sciences, Beijing, China ~~ $^{3}$Kingsoft Cloud, Beijing, China\\
	$^{4}$Peng Cheng Laboratory, Shenzhen, China ~~  $^{5}$Cloud BU, Huawei Technologies, Shenzhen, China.\\% , UCAS, Beijing, China.\\
	%$^{5}$Peng Cheng Laboratory, Shenzhen, China\\
	%$^{4}$Key Laboratory of Intell. Info. Process. (IIP), Inst. of Computi. Tech., CAS, China.\\
	{\tt\small hanxinzhe17@mails.ucas.ac.cn, wangshuhui@ict.ac.cn, suchi@kingsoft.com}\\
	{\tt\small qmhuang@ucas.ac.cn, tian.qi1@huawei.com}
}

%% omit the following lines up until the closing ``}''.
%% Additional authors and addresses can be added with ``\and'',
%% just like the second author.
%% To save space, use either the email address or home page, not both
%\and

\maketitle
\ificcvfinal\thispagestyle{empty}\fi

%%%%%%%%% ABSTRACT
\begin{abstract}
   Language bias is a critical issue in Visual Question Answering (VQA), where models often exploit dataset biases for the final decision without considering the image information. As a result, they suffer from performance drop on out-of-distribution data and inadequate visual explanation.
   Based on experimental analysis for existing robust VQA methods, we stress the language bias in VQA that comes from two aspects, {\it i.e.}, distribution bias and shortcut bias.
   We further propose a new de-bias framework, Greedy Gradient Ensemble (GGE), which combines multiple biased models for unbiased base model learning. With the greedy strategy, GGE forces the biased models to over-fit the biased data distribution in priority, thus makes the base model pay more attention to examples that are hard to solve by biased models. 
   The experiments demonstrate that our method makes better use of visual information and achieves state-of-the-art performance on diagnosing dataset VQA-CP without using extra annotations.
\end{abstract}

%%%%%%%%% BODY TEXT
\vspace{-1em}
\section{Introduction}
\begin{figure}[t]
	\begin{center}
		\subfigure[Distribution Bias]{
			\label{subfig:db}
			\includegraphics[width=0.9\linewidth]{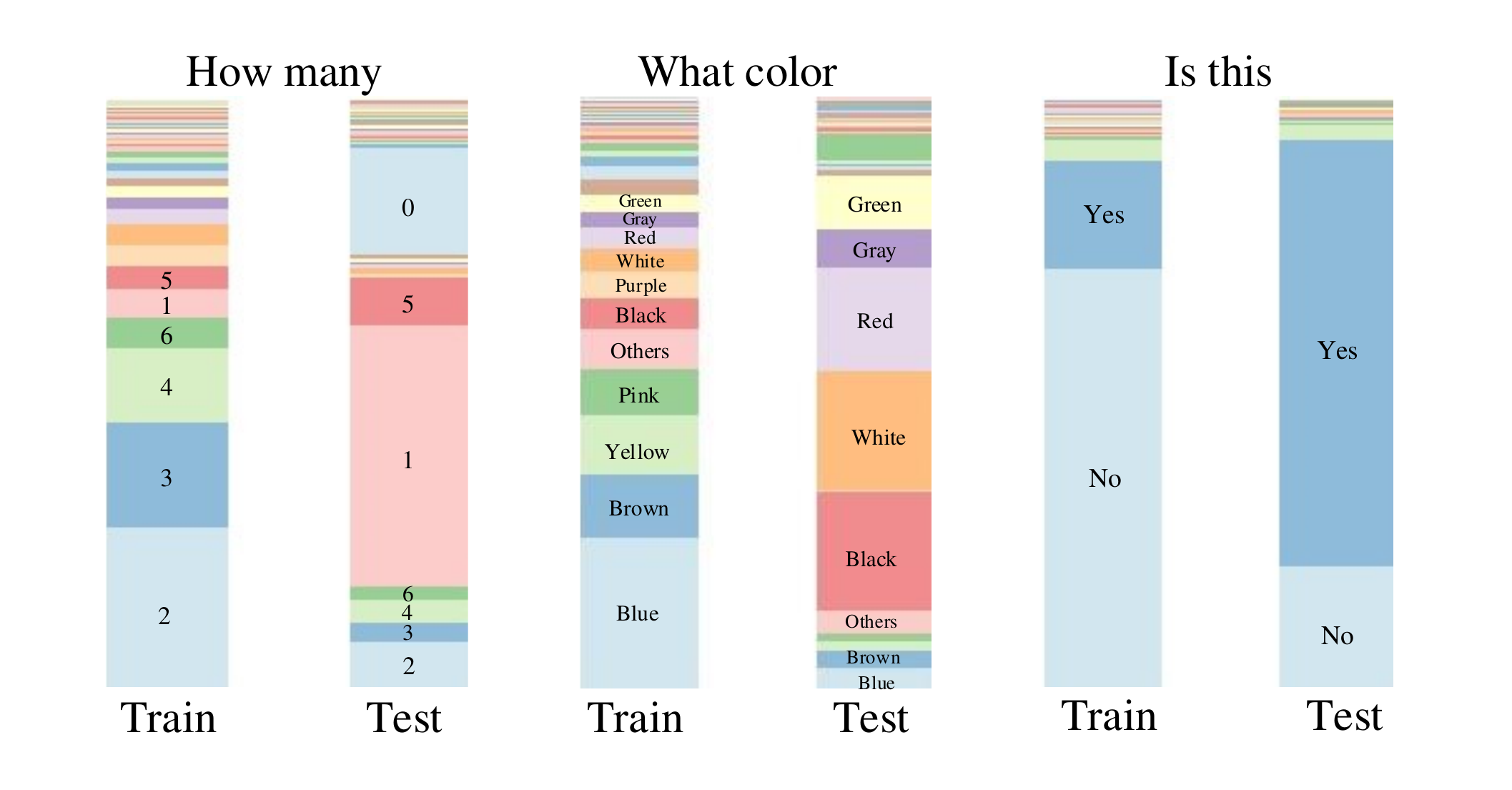}
		}\\
	\vspace{-0.5em}
		\subfigure[Shortcut Bias]{
			\label{subfig:lc}
			\includegraphics[width=\linewidth]{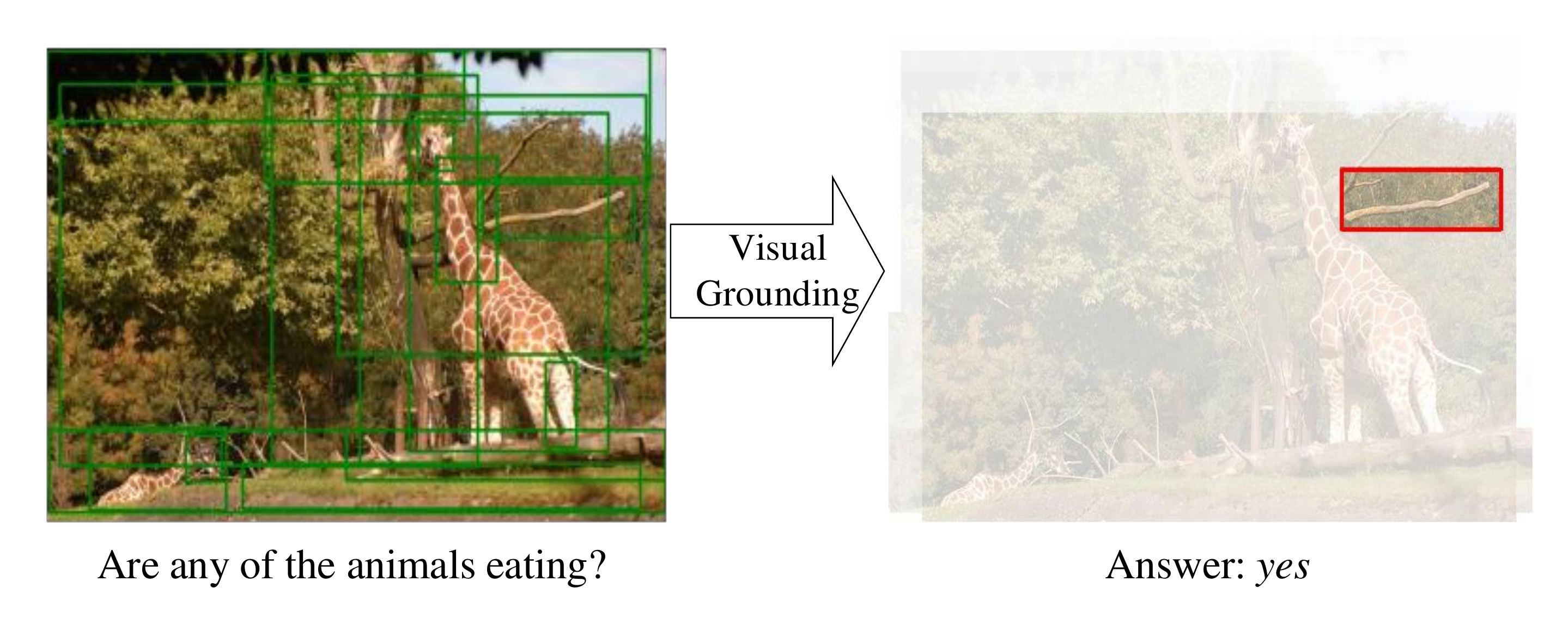}
		}
	\\	
	\end{center}
\vspace{-1em}
	\caption{Two aspects of language bias in VQA. 
	{\bf (a) Distribution Bias:} The answer distribution for certain question type is significantly long-tailed.
	{\bf (b) Shortcut Bias:} The correct answers produced by the model may rely on the question-answer shortcut rather than proper visual grounding.
	}
	\label{fig:2_bias}
\vspace{-1em}
\end{figure}

Visual Question Answering (VQA) is a challenging task that requires both language-aware reasoning and image understanding. With advances in deep learning, neural networks~\cite{2018film,2019murel,gao2019multi,2017n2nmn,2018stacknmn,2019lcgn,2019nscl,han2020interpretable} that model the correlations between vision and language have shown remarkable results on large-scale benchmark datasets~\cite{2015vqa,2017mfh,2017clevr,2019gqa}.

However, recent studies have demonstrated that most VQA methods tend to rely on existing idiosyncratic biases in the datasets~\cite{2017mfh,2017analysis,2016yin}. 
They often leverage superficial correlations between questions and answers to train the model without considering exact vision information. For example, a model may blindly answer ``tennis" for the question ``What sports ..." just based on the most common textual QA pairs in the train set. Unfortunately, models exploiting statistical shortcuts
%\footnote{Models use some unexpected statistical correlations for prediction.} 
during training often show poor generalization ability to out-of-domain data, and hardly provide proper visual evidence for a certain answer.%, especially if the dataset are carefully designed to limit the spurious cues like VQA-CP~\cite{2018vqacp}. 

%Different from long-tailed problem discussed in computer vision~\cite{2019large,2020bbn,2020decoupling,2019repair} and natural language processing~\cite{2018much,2018annotation,2019right,2020end}, bias in VQA is more complicated.
Currently, the prevailing solutions for this problem can be categorized into ensemble-based~\cite{2018overcoming,2019rubi,2019don}, grounding-based~\cite{2019hint,2019SCR,2020DLP} and counterfactual-based~\cite{2020counterfactual}. Similar to re-weighting and re-sampling strategies in traditional long-tailed classification~\cite{2020bbn,2020decoupling,2018annotation,2019right}, ensemble-based methods re-weight the samples by the question-only branch. %that a model is trained to be independent of.%,  
Grounding-based models stress a better use of image information according to human-annotated visual explanation~\cite{2017hat,2018vqax}. Newly proposed counterfactual-based methods~\cite{2020counterfactual,2020cf-vqa} further combine these two lines of work and achieve better performance.

Nevertheless, it has been shown that existing methods have not fully leveraged both vision and language information.
For example, Shrestha \etal~\cite{2020negative} argue that improved accuracy in grounding-based methods~\cite{2019hint,2019SCR} does not actually emerge from proper visual grounding but some unknown regularization effects. 
Similar to \cite{2020negative}, we further analyse all the three categories of existing work by control experiments in Section~3.2. 
We found that language bias in VQA is actually two-fold: (a) the statistical distribution gap between train and test, \ie, \emph{distribution bias} shown in Figure~\ref{subfig:db}, and (b) the semantic correlation between specific QA pairs, \ie, \emph{shortcut bias} shown in Figure~\ref{subfig:lc}. 
Although long-tailed distribution in train set is usually considered to be one of the factors that increase shortcut bias, we experimentally demonstrate that they are actually two aspects of the language bias. Grounding supervision in \cite{2019hint} or ensemble regularization in \cite{2019rubi,2019don} does not necessarily force the model to focus on visual information as expected. To encourage the model to pay attention to the images, we need to explicitly model bothbiases and reduce them step by step. 

%Similar to \cite{2020negative}, we further analyse all the three categories of existing works by control experiments.
%We found that language bias in VQA is actually from two folds: \emph{distribution bias} and \emph{shortcut bias}. Distribution bias is the statistical distribution gap between train and test as shown in Figure~\ref{subfig:db}. Models tend to over-fit the long-tailed label distribution in train set. Shortcut bias is the semantic correlation between specific QA pairs. Models may make predictions based on ``sufficiency" of language information regardless of the image as shown in Figure~\ref{subfig:lc}.
%Although long-tailed distribution in train set may increase the shortcut bias, we experimentally demonstrate that they are still two different aspects of language bias for VQA. Modelling such two entangled biases with just a question-only branch as in~\cite{2019rubi} leads to unsatisfactory model performance and inadequate visual interpretability. 

Inspired by our empirical findings, we propose Greedy Gradient Ensemble (GGE), a model-agnostic debias framework that ensembles biased models and the base model like gradient descent in functional space. The key idea of our method is to make use of the over-fitting phenomenon in deep learning. The biased part of data is greedily over-fitted by biased features, as a result, the expected base model can be learned with more ideal data distribution and focus on examples that are hard to solve with biased models. 

In the experiments, variants of GGE models are provided in ablation study, which demonstrates the generalization ability of our method and further supports our claim that distribution bias and shortcut bias are complementary in VQA. To verify if a model can really use visual information for the answer decision, we further study the language bias in VQA from a visual modelling perspective. Quantitative and qualitative evaluations show that GGE can provide better visual evidence accompanied with predictions.

%We further study the language bias in VQA by visual modelling perspective.
%We believe a model that can really make use of image information should not only provide the right answers based on the correct visual information, but also wrong answers due to wrong evidences as well. 
%To quantitatively evaluate the contribution from images, we propose a new metric called Correct Grounding Difference (CGD), which measures the score of attention agreement with human explanation. 
%As shown in the experiments, many models with high Accuracy on VQA-CP perform poorly on CGD. Even with remarkable Accuracy, the contribution from image to the final answer prediction may still be limited. GGE force the base model to pay more attention to samples that is hard to answer by language bias and achieve the highest CGD. As shown in Figure~\ref{fig:vis}, our method can also provide better visual evidences accompanied with correct predictions.

The major contributions are:
\begin{itemize}
	\item We provide analysis for the language bias in VQA task and decompose the language bias into distribution bias and shortcut bias.	
	
	\item We propose a new model-agnostic de-bias framework Greedy Gradient Ensemble (GGE), which sequentially ensembles biased models for robust VQA.
	
	\item On VQA-CP, our method makes better use of visual information and achieves state-of-the-art performance, with $17.34\%$ gain against simple UpDn baseline without extra annotations. Code is available at \url{https://github.com/GeraldHan/GGE}.

\end{itemize}

%inspired by ensemble strategy in boosting~\cite{2000gradient,2016xgboost}, we propose 
%Different from boosting methods where base classifiers should be weak enough to avoid over-fitting, Greedy-VQA take advantage of the over-fitting of certain biased classifier and remove the distribution bias and language bias step by step. 

\section{Related work}
\subsection{De-bias with dataset construction}

The most straightforward way to remove the dataset bias is to construct a balanced dataset. For instance, Zhang \etal~\cite{2016yin} collect complementary abstract scenes with opposite answers for all binary questions.
Similarly, VQA v2~\cite{2017mfh} is introduced to weaken language priors in the VQA v1 dataset~\cite{2015vqa} by adding similar images with different answers for each question. Agrawal \etal~\cite{2018vqacp} introduce a diagnosing VQA dataset under Changing Prior (VQA-CP) constructed with different answer distributions between the train and test splits. %Hence models that rely heavily on the language priors in the train set will show poor generalization while testing. 
Most of the models that perform well on VQA v2 significantly drop on VQA-CP in Accuracy. %This indicates that it is hard to remove bias by dataset construction protocols in cross-modal tasks like VQA.

\subsection{De-bias with model design}
Collecting new large-scale datasets is costly. It is crucial to develop models that are robust to biases. Along with VQA-CP dataset~\cite{2018vqacp}, Agrawal \etal propose GVQA model that disentangles the visual concept recognition from the answer space prediction. LDP~\cite{2020DLP} and GVQE~\cite{2020reducing} exploit different information in questions for better question representation. 
These models require a pre-defined question parser, making them hard to implement.
 
Another line of work starts from visual grounding. Early works~\cite{2017exploring,2019interpretable} directly apply human grounding~\cite{2017hat,2018vqax} as supervision to attention maps, but the improvement is limited. HINT~\cite{2019hint} and SCR~\cite{2019SCR} change supervised attention maps to Grad-CAM, which directly encourages the contribution of each object to be consistent with human annotations. Recent work~\cite{2020negative} experimentally challenges the effectiveness of visual grounding in \cite{2019hint,2019SCR}, but does not find a good way to test if systems are actually visually grounded.

The most effective solution so far is ensemble-based, which formulates a question-only branch as explicit modelling for language bias. 
Ramakrishnan \etal~\cite{2018overcoming} introduce an adversarial regularization to remove answer discriminative feature from the questions. RUBi~\cite{2019rubi}, LMH~\cite{2019don} and PoE~\cite{2020end} re-weight samples based on the question-only prediction. Niu \etal~\cite{2020cf-vqa} further improve ensemble strategies from a causal-effect perspective. CSS~\cite{2020counterfactual} combines grounding-based and ensemble-based methods with counterfactual samples synthesizing. 
%It is actually a kind of data augmentation for more complementary samples. 
Gat \etal~\cite{2020mfe} introduce a regularization by maximizing functional entropies (MFE), which forces the model to use multiple sources of information in multi-modal tasks. Nam \etal~\cite{2020lff} propose a general framework LfF, which trains the de-biased classifier from a biased classifier. Compared to our work, they mainly focus on single-modality classification problems and their General Cross-Entropy (GCE) re-weighting strategy is less flexible, which relies on hyper-parameter in GCE and can only handle one pair of attributes in de-bias learning. 
%Recent work~\cite{2020negative} experimentally challenge the effectiveness of visual grounding in grounding-based methods, but do not find a good way to test if systems are actually visually grounded. In this work, we provide more deep-in analysis from a de-bias method perspective.

%\subsection{Boosting}
%Boosting is a widely used ensemble strategy for solving classification problems. Since the seminal works of Freund~\cite{1995boosting}, Schapire~\cite{schapire1990strength}, a number of practical algorithms such as AdaBoost~\cite{1996adaboost}, gradient boosting~\cite{2000gradient}, XGBoost~\cite{2016xgboost}, have been proposed. The key idea of boosting is to greedily combine multiple weak classifiers with high bias but low variance to produce a strong classifier with low bias low variance. Each base learner has to be weak enough, otherwise the first few classifier will easily over-fit to the training data~\cite{2006some}. However, the fitting ability of neural networks are too strong to be ``low variance" for boosting strategy. In this paper, we take use of the over-fitting phenomenon, making biased weak feature to over-fit bias distribution, thus remove distribution bias in VQA datasets.

%Long-tailed classification problem has become a hot spot in both computer vision~\cite{2019large,2020bbn,2020decoupling,2019repair} and natural language processing~\cite{2018much,2018annotation,2019right,2020end} community.

\section{Revisiting Language Bias in VQA}
\subsection{Problem Definition}
For base model, we consider the common formulation of VQA task as a multi-class classification problem. Given a dataset $\mathcal{D} = \{v_i, q_i, a_i \}_{i=1}^N$ consisting of an image $v_i \in \mathcal{V}$, a question $q_i \in \mathcal{Q}$ and a labelled answer $a_i \in \mathcal{A}$, we need to optimize a mapping $f_{VQ}: V \times Q \rightarrow \mathbb{R}^C$ which produces a distribution over the $C$ answer candidates. Without loss of generality, the function is composed as following:
\begin{equation}\label{baseline}
\tilde{a}_i = f_\theta(v_i, q_i) = c\left(m\left(e_v(v_i), e_q(q_i) \right)\right),
\end{equation}
where $e_v: V \rightarrow \mathbb{R}^{n_v \times d_v}$ is an image encoder, $e_q: Q \rightarrow \mathbb{R}^{n_q \times d_q}$ is a question encoder, $m(.)$ denotes the multi-modal fusion or reasoning module, and $c(.)$ is the multi-layer perception classifier. The output is a vector $\tilde{a} \in \mathbb{R}^C$ indicating the probability belonging to each answer candidate. 
%Most VQA methods~\cite{2018bottomup,2018BAN,2019lcgn,2019murel} for VQA v2~\cite{2017mfh} can be modelled as the above formulation. 
%Commonly used UpDn~\cite{2018bottomup} is adopted as the base model in the following experiments unless specifically clarified.

\subsection{Experimental Analysis for Language Bias}
In recent work, Shrestha \etal~\cite{2020negative} experimentally challenge the way grounding-based methods~\cite{2019SCR,2019hint} work on VQA-CP~\cite{2018vqacp}. However, they did not provide insights into the language bias itself.
In this section, from a new de-bias method perspective, we provide three control experiments for baseline model UpDn~\cite{2018bottomup}, grounding-based method HINT~\cite{2019hint}, ensemble-based method RUBi~\cite{2019rubi} LMH~\cite{2019don} and counterfactual-based method CSS~\cite{2020counterfactual} on VQA-CP and VQA v2 to discuss the language bias in VQA. 

{\bf Inverse Grounding Annotation}. To analyse the contribution of visual-grounding, we first experiment with HINT and CSS-V that use human attention as extra information. Following \cite{2020negative}, we change human-annotated region importance scores~\cite{2017hat} $S_h$ to irrelevant grounding $S_h' = 1 - S_h$. As shown in Table~\ref{tab:analysis}, the performance of HINT$_{inv}$ and CSS-V$_{inv}$ is almost the same as the original models. This indicates that the Accuracy gains are not necessarily from looking at relevant regions~\cite{bai2020attention}. Although the models correctly answer some hard questions, they still make predictions based on language information regardless of images. We refer to this unexpected solution as ``inverse language bias".

{\bf Vision-only Model}. 
The second experiment aims to analyse the function of the ensemble branch in RUBi and LMH. For the base model,
we only feed the vision feature without multi-modal fusion to the answer classifier:
\begin{equation}\label{vonly}
\tilde{a}_i = c\left(e_v(v_i) \right).
\end{equation}
%Question-only branch in LMH only works as a regularization in train and is removed in test. 
There is no question information for classification in base model, and thus obviously no shortcut between QA pairs to reduce. As shown in Table~\ref{tab:analysis}, RUBi$_{vo}$ degrades a lot, but LMH$_{vo}$ still surpasses UpDn$_{vo}$ by a large margin in Accuracy. 
Apart from restraining shortcuts between question-answer pairs, we think the improved Accuracy in LMH mainly comes from penalizing the most common answers in the train set, which leads to a more balanced classifier according to inverse distribution. This means the distribution bias in LMH plays a different role compared with the question shortcut in RUBi.
%
%this is because ensemble-based methods use the answer distribution discrepancy between train and test set (distribution bias). Apart from restraining shortcut bias, the improved Accuracy also comes from a balanced the classifier according to ``inverse" distribution bias.

{\bf Inverse Supervision for Balanced Classifier}. To directly verify if such ``inverse distribution bias" can improve Accuracy, inspired by the two-round training in CSS~\cite{2020counterfactual}, we design a simple ``inverse supervision" strategy. For each iteration, the parameters are updated two rounds with different supervisions. In the first round, we train the model supervised by ground-truth label $\mathcal{A}$ and get the prediction $P(a)$. The top-$N$ answers with the highest predicted probabilities are selected as $\mathbf{a^+}$. In the second-round training, the label is defined as $\mathcal{\hat{A}} = \{a_i | a_i \in \mathcal{A}, a_i \notin \mathbf{a^+} \}$. 
This strategy is actually a simplified version of CSS~\cite{2020counterfactual} without object/question masks. In this way, the model continuously penalizes the most confident answers in the first round training, thus formulates a more balanced classifier according to inverse distribution bias. The Accuracy improvement in UpDn$_{vo,is}$ reveals the existence of distribution bias. The result of RUBi$_{is}$ further indicates that distribution bias and shortcut bias are complementary. LMH$_{is}$ is even comparable to CSS that uses extra annotations. However, this method leads to catastrophic degradation on the in-distribution dataset VQA v2 as shown in Table~\ref{tab:analysis}. %We provide detailed results in Section~D in the Supplementary.
\begin{table}[t]
	\centering
	\caption{Experimental analysis for representative methods on VQA-CP and VQA v2. Footnote $inv$ stands for Inverse Grounding Annotation, 
		%$gt-att$ denote train with Human Annotated Attention, 
		$vo$ for Vision-only,
		and $is$ for Inverse-Supervision.}
	\setlength{\tabcolsep}{6.5mm}
	\begin{tabular}{l|cccc}
		\hline
		Method            & VQA-CP & VQA 2.0  \\ \hline
		UpDn~\cite{2018bottomup}        &  39.89 & 63.79     \\
		HINT~\cite{2019hint}            &  47.50 & 63.38    \\
		RUBi~\cite{2019rubi}            &  45.42 & 58.19   \\
		LMH~\cite{2019don}              &  52.73 & 56.35     \\
		CSS~\cite{2020counterfactual}   &  58.11 & 53.15     \\ \hline
		HINT$_{inv}$  &  47.20  & 60.33    \\
		CSS-V$_{inv}$ &  58.05  & 54.39    \\ \hline
		UpDn$_{vo}$   & 33.18   & 45.67    \\
		RUBi$_{vo}$   & 23.53   & 46.11  \\ 
		LMH$_{vo}$    & 43.68       & 27.18 \\
		\hline
		%UpDn$_{gt-att}$   &  34.33      & -     & -     & -    \\ \hline
		UpDn$_{vo,is}$ & 39.44  &  40.03      \\
		UpDn$_{is}$ &  42.12    & 60.85   \\
		RUBi$_{is}$ &  48.42    & 59.10   \\
		LMH$_{is}$  &  58.12    & 43.29   \\ 
		\hline
	\end{tabular}
	\label{tab:analysis}
	\vspace{-2em}
\end{table}

According to the above experiments, we obtain the following insights:
1) Good Accuracy can not guarantee that the system is really visually grounded for answer classification. Grounding supervision or question-only regularization may encourage models to make use of inverse language bias rather than better visual information for higher Accuracy.
2) Distribution bias and shortcut bias are complementary aspects of language bias in VQA. A single ensemble branch is unable to model such two types of biases.

\section{Method}
Based on the above findings, we propose GGE, a new model-agnostic de-bias learning paradigm, which removes distribution bias and shortcut bias step by step, thus forces the model to focus on images.
%Different from previous ensemble-based de-bias methods~\cite{2019don,2019rubi,2020end} specially designed for one particular bias branch, GGE can adapt to any number of bias hypotheses as long as the bias features can be disentangled with prior knowledge.

\subsection{Greedy Gradient Ensemble}
Let $(X,Y)$ denote the train set, where $X$ is the space of observations, and $Y$ is the space of labels. Following previous VQA methods, we mainly consider the classification problem with binary cross-entropy (BCE) loss
\begin{equation}\label{bce}
\mathcal{L}(P, Y) = -\sum_{i=1}^{C} y_i \log(p_i) +  (1- y_i) \log(1- p_i),
\end{equation}
where $C$ denotes the number of classes. $p_i = \sigma(z_i)$ where $z_i$ is the predicted logit for class $i$  and $\sigma(.)$ is the sigmoid function. Baseline methods directly minimize the loss between the prediction $f(X;\theta)$ and label $Y$
\begin{equation}\label{base}
\min_\theta \mathcal{L}\left(\sigma(f(X; \theta)), Y  \right).
\end{equation}
Since $f(.)$ is over-parametrized DNNs, the model is easy to over-fit the dataset biases and suffers from poor generalization ability.

For our method, we make use of this kind of over-fitting in deep learning. Assume $\mathcal{B} = \{B_1, B_2, \dots, B_M\}$ to be a set of bias features that can be extracted based on prior knowledge. This time we fit the ensemble of bias models and base model to label $Y$
\begin{equation}\label{base}
\min_{\phi,\theta} \mathcal{L}\left( \sigma\left(f(X; \theta) + \sum_{i=1}^{M}h_i(B_i; \phi_i) \right), Y  \right),
\end{equation}
where $h_i(.)$ is a biased model for certain biased feature. 
Ideally, we hope the biased part of data is only over-fitted by the bias models, thus the base model can be learned with unbiased data distribution. To achieve this goal, we propose GGE in which biased models have a higher priority to over-fit the dataset with greedy strategy. 

Viewing in the functional space, suppose we have $\mathcal{H}_m = \sum_{i=1}^{m}h_i(B_i)$ and we wish to find $h_{m+1}(B_{m+1})$ added to $\mathcal{H}_m$ so that the loss $\mathcal{L}\left(\sigma(\mathcal{H}_m + h_{m+1}(B_{m+1})), Y  \right)$ decreases. In theory, the desired direction of $h_{m+1}$ is the negative derivative of $\mathcal{L}$ at $\mathcal{H}_m$, where
\begin{equation}\label{direction}
-\nabla \mathcal{L}(\mathcal{H}_{m,i}) := \frac{\partial \mathcal{L}\left(\sigma(\mathcal{H}_m), Y  \right)}{\partial \mathcal{H}_{m,i}} = 2y_{m,i} \sigma \left(-2 y_{m,i} \mathcal{H}_{m,i} \right).
\end{equation}
For a classification problem, we only care about the probability for class $i$: $\sigma(f_i(x)) \in \{0,1\} $. Therefore, we treat the negative gradients as pseudo labels for classification and optimize the new model $h_{m+1}(B_{m+1})$ with BCE loss:
\begin{equation}\label{gradient}
%\min_{\phi_{m+1}} \mathcal{L} \left(\sigma(h_{m+1}(B_{m+1};\phi_{m+1})), -\nabla \mathcal{L}(\mathcal{H}_m)\right)
L_{m+1} = \mathcal{L} \left(\sigma(h_{m+1}(B_{m+1};\phi_{m+1})), -\nabla \mathcal{L}(\mathcal{H}_m)\right).
\end{equation}
%Intuitively, it makes $h_{m+1}$ pay more attention to the samples that are hard to solve by previous ensemble biased classifiers $\mathcal{H}_m$, similar to adaptive instance re-weighting in \cite{2020bbn,2020decoupling,2018annotation,2019right}.

After integrating all biased models, the expected base model $f$ is optimized with
\begin{equation}\label{fx}
%\min_\theta \mathcal{L} \left(\sigma(f(X;\theta)), -\nabla \mathcal{L}(\mathcal{H}_M) \right)
L_b(\theta) = \mathcal{L} \left(\sigma(f(X;\theta)), -\nabla \mathcal{L}(\mathcal{H}_M) \right).
\end{equation}
In the test stage, we only use the base model for predictions. 

More intuitively, for a sample that is easy to fit by biased models, the negative gradient of its loss $-\nabla \mathcal{L}(\mathcal{H}_M)$ (\ie, the pseudo label for the base model) will become relatively small. $f(X;\theta)$ will pay more attention to samples that are hard to solve by previous ensemble biased classifiers $\mathcal{H}_M$.

In order to make the above paradigm adaptive to Batch Stochastic Gradient Decent (Batch SGD), we implement two optimization schedules GGE-iteration and GGE-together, as shown in Algorithm~1 and Algorithm~2 in Supplementary. GGE-tog jointly optimizes biased models and the base model with 
\begin{equation}
L(\Theta) = L_b(\theta) + \sum_{m=1}^{M} L_m(\phi_{m}).
\end{equation}
For GGE-iter, each model is iteratively updated within a certain data-batch iteration. 
More details for GGE are provided in Section A in Supplementary.

%In theory, one can also change the gradient with other metrics that can measure the distance between labels and predictions.

\begin{figure*}[t]
	\begin{center}
		\subfigure[Baseline]{
			\label{subfig:baseline}
			\includegraphics[width=0.39\linewidth]{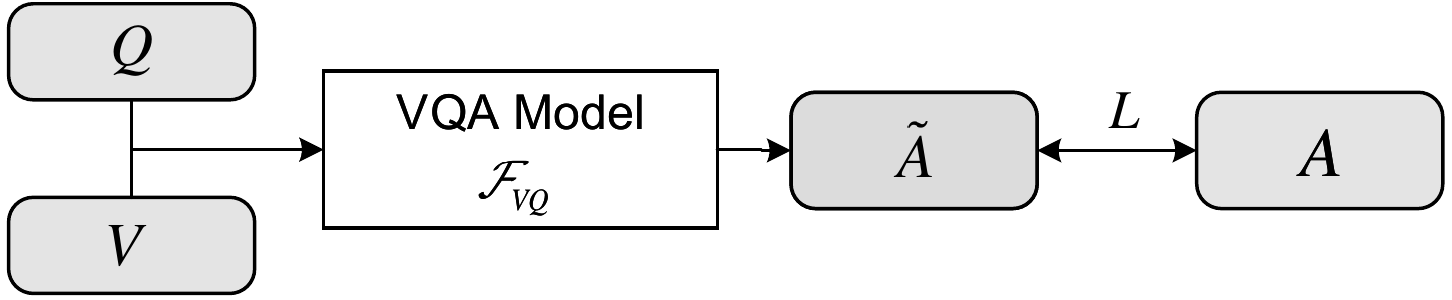}
		}
		\hspace{2em}
		\subfigure[GGE-D]{
			\label{subfig:dist_bias}
			\includegraphics[width=0.4\linewidth]{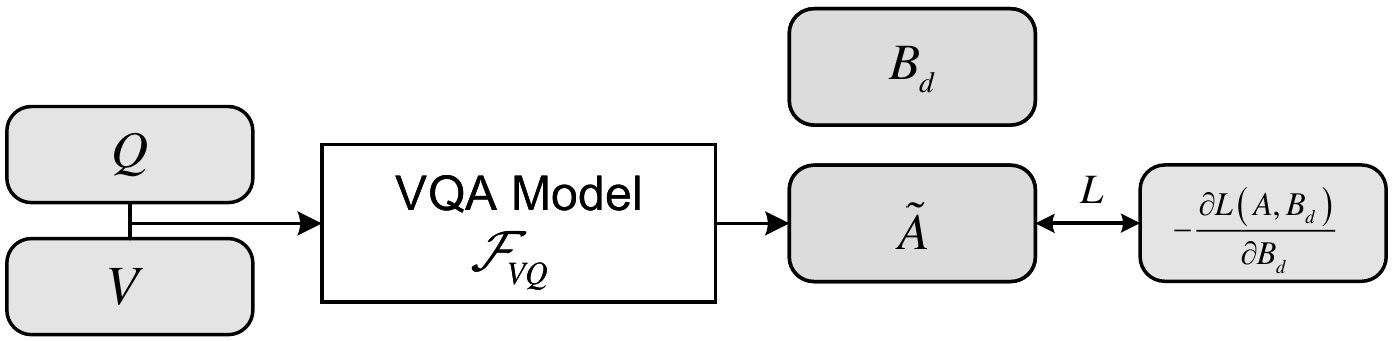}
		}\\
		\subfigure[GGE-Q]{
			\label{subfig:q_bias}
			\includegraphics[width=0.4\linewidth]{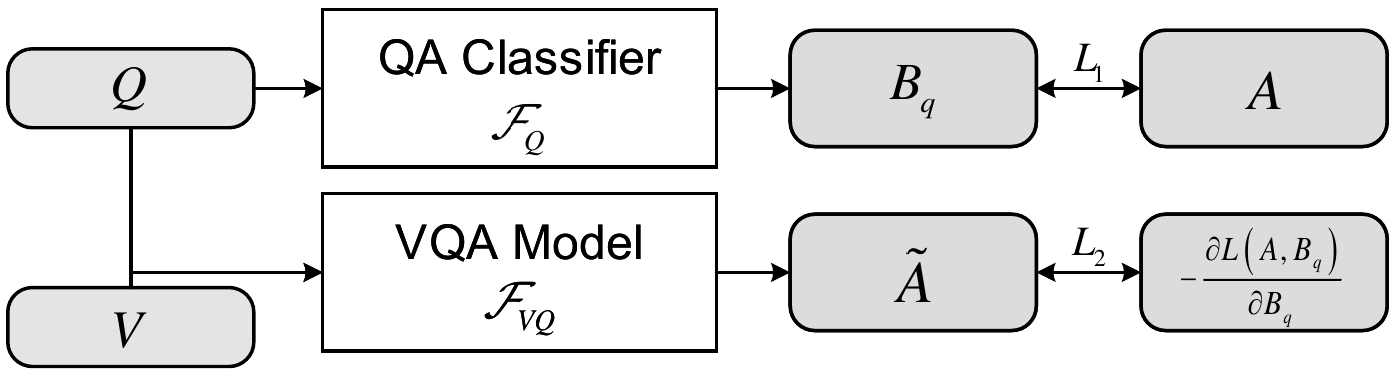}
		}
		\hspace{2em}
		\subfigure[GGE-DQ]{
			\label{subfig:dist_q_bias}
			\includegraphics[width=0.4\linewidth]{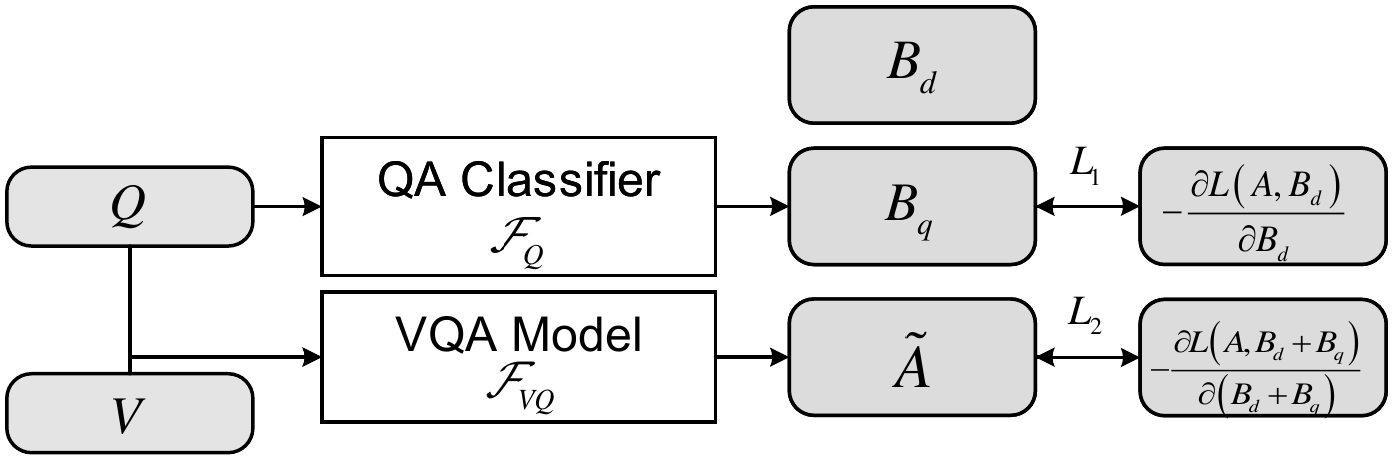}
		}
	\end{center}
\vspace{-1em}
	\caption{{\bf Different versions of GGE}. $V,Q$ and $\tilde{A}$ denote image, question, and answer prediction respectively. $A$ is the human-annotated labels. $B_d$ and $B_q$ indicate the prediction from distribution bias and question shortcut bias respectively.} %In our experiments, we use UpDn~\cite{2018bottomup} as our base VQA Model.}
	\label{fig:model}
	\vspace{-1.5em}
\end{figure*}
\subsection{GGE Implementation for Robust VQA} 
Following analysis in Section~3, we define two biased features for VQA, \ie, distribution bias and shortcut bias.

{\bf Distribution Bias}. %As shown in Figure~\ref{subfig:db}, the answer distribution for a certain question type is quite different between train and test. 
We define the distribution bias as answer distribution in the train set conditioned on question types
\begin{equation}\label{db}
B_{d}^i = p(a_i | t_i),
\end{equation}
where $t_i$ denotes the type of question $q_i$. The reason for counting samples conditioned on question types is to maintain type information when reducing distribution bias. Question type information can only be obtained from the questions rather than the images, which does not belong to the language bias to be reduced.

{\bf Shortcut Bias}. Shortcut bias is the semantic correlation between specific QA pairs. Similar to \cite{2019rubi}, we compose the question shortcut bias as a question-only branch
\begin{equation}\label{sb}
B_{q}^i = c_q \left(e_q(q_i)  \right),
\end{equation}
where $c_q: Q \rightarrow \mathbb{R}^C$.

To verify our claim that distribution bias and shortcut bias are complementary, we design three versions of GGE for ensembles of different language biases.

{\bf GGE-D} only models distribution bias for ensemble, shown in Figure~\ref{subfig:dist_bias}. The loss for the base model is 
\begin{equation}\label{gge_b}
L = \mathcal{L}(\sigma(\tilde{A}), -\nabla \mathcal{L}(B_d, A) ),
\end{equation}
where $\tilde{A}$ is the predictions, and $A$ is the labelled answers.

{\bf GGE-Q} only uses a question-only branch for shortcut bias. As shown in Figure~\ref{subfig:q_bias}, we first optimize the question-only branch with labelled answers
\begin{equation}\label{gge_q}
L_1 = \mathcal{L}(\sigma(B_q), A).
\end{equation}
The loss for base model is 
\begin{equation}\label{gge_q2}
L_2 = \mathcal{L}(\sigma(\tilde{A}), -\nabla \mathcal{L}(\sigma(B_q), A) ).
\end{equation}

{\bf GGE-DQ} uses both distribution bias and question shortcut bias. As shown in Figure~\ref{subfig:dist_q_bias}, the loss for $B_q$ is 
\begin{equation}\label{l_q}
L_1 = \mathcal{L}(\sigma(B_q), -\nabla \mathcal{L}(B_d, A) ).
\end{equation}
The loss for base model is
\begin{equation}\label{l_dq}
L_2 = \mathcal{L}(\sigma(\tilde{A}), -\nabla \mathcal{L}(\sigma(B_q) + B_d , A) ).
\end{equation}
We test both GGE-iter or GGE-tog for $L_1$ and $L_2$.

\subsection{Connection to Boosting}
Boosting~\cite{1995boosting,schapire1990strength,schapire1990strength,2016xgboost} is a widely used ensemble strategy for classification problems. 
%Since the seminal works of Freund~\cite{1995boosting}, Schapire~\cite{schapire1990strength}, a number of practical algorithms such as AdaBoost~\cite{1996adaboost}, gradient boosting~\cite{2000gradient}, XGBoost~\cite{2016xgboost}, have been proposed. 
The key idea of boosting is to combine multiple weak classifiers with high bias but low variance to produce a strong classifier with low bias and low variance. Each base learner has to be weak enough, otherwise, the first few classifiers will easily over-fit to the training data~\cite{2006some}. However, the neural networks' fitting ability is too strong to be ``high bias" and ``low variance" for boosting strategy, making it hard to use deep models as weak learners. 
In this paper, our method exploits this over-fitting phenomenon, making biased weak features to over-fit the bias distribution. In the test stage, we only use the base model trained with the gradient of biased models, thus removing language bias in VQA. 

On the other hand, the idea of approximating negative gradients is very similar to Gradient Boost~\cite{2000gradient}.
However, Gradient Boost has to greedily learn weak learners in turn. This will be costly for complicated neural networks via back-propagation. We design two strategies, GGE-iteration and GGE-together, in which the learners are updated along with Batch SGD.

\begin{table*}[t]
	\small
	\centering
	\setlength{\tabcolsep}{2.9mm}
	\caption{Experimental results on VQA-CP v2 test set and VQA v2 val set of state-of-the-art methods. \textbf{\underline{Best}} and \textbf{second} performance are highlighted in each column. 
		Methods with * use extra annotations (\eg, human attention (HAT), explanations (VQA-X), or object label information). Methods with CGD are our reimplementation using released codes. Other results are reported in the original papers. }
	\label{tab:sota}
	\begin{tabular}{lccccccccccc}
		\hline
		\multirow{2}{*}{Method} & \multirow{2}{*}{Base} & \multicolumn{5}{c}{VQA-CP test} &  & \multicolumn{4}{c}{VQA v2 val} \\ \cline{3-7} \cline{9-12} 
		&                       & All & Y/N & Num. & Others & CGD &  & All  & Y/N  & Num. & Others \\ \hline
		GVQA~\cite{2018vqacp}   &  -     & 31.30 & 57.99 & 13.68 & 22.14 & -     & & 48.24 & 72.03 & 31.17 & 34.65 \\
		UpDn~\cite{2018bottomup}&  -     & 39.89 & 43.01 & 12.07 & 45.82 & 3.91  & & \textbf{63.79} & 80.94 & 42.51 & \textbf{\underline{55.78}} \\
		S-MRL~\cite{2019rubi}   &  -     & 38.46 & 42.85 & 12.81 & 43.20 & -     & & 63.10 & -     & -     & -     \\ \hline
		HINT*~\cite{2019hint}   & UpDn   & 47.50 & 67.21 & 10.67 & 46.80 & 10.34 & & 63.38 & \textbf{81.18} & 42.14 & \textbf{55.66} \\
		SCR*~\cite{2019SCR}     & UpDn   & 49.45 & 72.36 & 10.93 & 48.02 & -     & & 62.2  & 78.8  & 41.6  & 54.4  \\
		AdvReg.~\cite{2018overcoming}& UpDn   & 41.17 & 65.49 & 15.48 & 35.48 & -     & & 62.75 & 79.84 & 42.35 & 55.16     \\
		RUBi~\cite{2019rubi}    & UpDn   & 45.42 & 63.03 & 11.91 & 44.33 & 6.27  & & 58.19 & 63.04 & 41.00 & 54.43     \\
		LM~\cite{2019don}       & UpDn   & 48.78 & 70.37 & 14.24 & 46.42 & 11.33     & & 63.26 & 81.16 & 42.22 & 55.22 \\
		LMH~\cite{2019don}      & UpDn   & 52.73 & 72.95 & \textbf{31.90} & 47.79 & 10.60 & & 56.35 & 65.06 & 37.63 & 54.69 \\
		DLP~\cite{2020DLP}      & UpDn   & 48.87 & 70.99 & 18.72 & 45.57 & - & & 57.96 & 76.82 & 39.33 & 48.54  \\
		GVQE*~\cite{2020reducing}& UpDn   & 48.75 & -     & -     & -     & -     & & \textbf{\underline{64.04}} & -     & -     & -    \\
		CSS*~\cite{2020counterfactual}& UpDn   & 41.16 & 43.96 & 12.78 & 47.48 & 8.23  & & 59.21 & 72.97 & 40.00 & 55.13  \\
		CF-VQA(Sum)~\cite{2020cf-vqa}& UpDn   & 53.69 & \textbf{\underline{91.25}} & 12.80 & 45.23 & -     & & 63.65 & \textbf{\underline{82.63}} & \textbf{\underline{44.01}} & 54.38 \\ \hline
		GGE-DQ-iter (Ours)         & UpDn   & \textbf{57.12} & 87.35 & 26.16 & \textbf{\underline{49.77}} & \textbf{\underline{16.44}} & & 59.30 & 73.63 & 40.30 & 54.29  \\
		GGE-DQ-tog (Ours)          & UpDn   & \textbf{\underline{57.32}} & 87.04 & 27.75 & \textbf{49.59} & \textbf{15.27} & & 59.11 & 73.27 & 39.99 & 54.39 \\ \hline \hline
		RUBi~\cite{2019rubi}    & S-MRL  & 47.11 & 68.65 & 20.28 & 43.18 & -     & & 61.16 & -     & -     & -     \\	
		GVQE*~\cite{2020reducing}& S-MRL  & 50.11 & 66.35 & 27.08 & 46.77 & -     & & 63.18 & -   & -  &  -   \\
		CF-VQA(Sum)~\cite{2020cf-vqa}& S-MRL  & 54.95 & \textbf{90.56} & 21.88 & 45.36 & -     & & 60.76 & 81.11 & \textbf{43.48} & 49.58 \\ \hline \hline
		MFE~\cite{2020mfe}& LMH    & 54.55 & 74.03 &  \textbf{\underline{49.16}} & 45.82 & -  & & - & - & - & -   \\
		CSS*~\cite{2020counterfactual}& LMH    & 58.21 & 83.65 & 40.73 & 48.14 & 8.81  & & 53.15 & 61.20 & 37.65 & 53.36   \\
		\hline
	\end{tabular}
\vspace{-1em}
\end{table*}

\section{Experiments}
The experiments are conducted on both language-bias sensitive VQA-CP v2~\cite{2018vqacp} and standard VQA v2~\cite{2017mfh}. 
%For a fair comparison with other methods, we use UpDn~\cite{2018bottomup} as our baseline model. 
Considering there is no validation set for VQA-CP, we simply choose the model in the last training epoch for comparison in consequent experiments.
More implementation details can be found in Section C in the Supplementary.

\subsection{Evaluation Metrics}
For each model, we compare Accuracy, the standard VQA evaluation metric~\cite{2015vqa}. Moreover, a robust VQA model is expected to leverage both visual and language information, but good Accuracy is not enough to indicate the system is well visually grounded according to analysis in Sec.~3.

In \cite{2020negative}, a new metric Correctly Predicted but Improperly Grounded (CPIG) is proposed to quantitatively assess visual grounding in VQA. An instance is regarded as correctly grounded if the ground-truth regions for the right answer (\eg, HAT~\cite{2018vqax}) are within the model's top-$N$ most sensitive visual regions. 
For convenience, we define $1-CPIG$ as $CGR$ (Correct Grounding for Right prediction):
\begin{equation}\label{rr}
\%CGR = \frac{N_{\text{rg,rp}}}{N_{\text{rp}}} \times 100\%,
\end{equation}
where $N_{\text{rp}}$ is the total number of right predictions, $N_{\text{rg,rp}}$ is the number of instances that are correctly answered with correct visual grounding. However, similar to results in \cite{2020negative}, we find that CGR is not very discriminative across different methods as shown in Table~2 in Supplementary. The model with high CGR (\eg, UpDn) may not actually use enough visual information for classification. If a model locates the right object but still produces a wrong answer, it is a safe bet that it heavily relies on language bias instead of images for prediction.
To quantitatively assess whether a model uses visual information for answer decision, we introduce CGW (Correct Grounding but Wrong prediction):
\begin{equation}\label{rw}
\%CGW = \frac{N_{\text{rg, wp}}}{N_{\text{wp}}} \times 100\%,
\end{equation}
where $N_{\text{wp}}$ is the number of wrong predictions, and $N_{\text{rg,wp}}$ is the number of instances for which the model provides the right visual evidences but wrong prediction. 
Bad cases like example 2 and 3 from UpDn in Fig.~\ref{fig:vis} are ignored by CGR but can be identified by high CGW.

For clearer comparison, we denote the difference of CGR and CGW as CGD (Correct Grounding Difference):
\begin{equation}\label{cgd}
\%CGD = \%CGR - \%CGW.
\end{equation}
CGD \emph{only} evaluates whether the visual information is taken in answer decision, which is parallel with Accuracy. The key idea for CGD is that a model actually makes use of visual information should not only provide the right predictions based on the correct visual-groundings but also a wrong answer due to improper visual evidence as well. Detailed CGR and GCD for all experiments are provided in Table~2 in Supplementary. It shows that UpDn, HINT$_{inv}$ and CSS-V$_{inv}$ achieve comparable performance on Accuracy but significantly degrade on CGD. This meets our intuitive analysis that these methods do not fully exploit visual information for the answer decision.
Although the visual-grounding annotations are not so reliable for some instances\footnote{Not all examples in VQA v2 are annotated in VQAX~\cite{2017hat}. Moreover, visual grounding for some instances are hard to evaluate (\eg, questions that require global image information or without referring objects)},
CGD can offer statistically better distinction from the whole dataset level. More details for CGD are provided in Section~B in the Supplementary.

\subsection{Comparison with State-of-the-art Methods}
We compare our best performed model GGE-DQ with existing state-of-the-art bias reduction techniques, including visual grounding-based methods HINT~\cite{2019hint}, SCR~\cite{2019SCR}, ensemble-based methods AdvReg.~\cite{2018overcoming}, RUBi~\cite{2019rubi}, LM (LMH)~\cite{2019don}, MFE~\cite{2020mfe}, new question encoding-based methods GVQE~\cite{2020reducing}, DLP~\cite{2020cf-vqa}, counterfactual-based methods CF-VQA~\cite{2020cf-vqa}, CSS~\cite{2020counterfactual}
%\footnote{We do not highlight LMH-CSS in Table~\ref{tab:sota}, since it uses extra information and suffered from dramatic degradation on VQA v2.}
, and recent proposed regularization method MFE~\cite{2020mfe}. 

Experiments on VQA-CP test set aim to evaluate whether VQA models effectively reduce language bias. As shown in Table~\ref{tab:sota}, GGE-DQ achieves state-of-the-art performance without extra annotation. It outperforms the baseline model UpDn by 17\% higher in Accuracy and 13\% higher in CGD, which verifies the effectiveness of GGE on both answer classification and visual-grounding ability. 
Under the same base model UpDn, our method achieves the best performance in both Accuracy and CGD, with $\sim5\%$ gain comparing to all other methods, even competitive with methods that use stronger base models. 

For the comparison of question-type-wise results, incorporating GGE reduces the biases and improves the performance for all the question-types, especially the more challenging ``other" question type~\cite{teney2020value}. 
CF-VQA~\cite{2020cf-vqa} performs the best in Y/N, but worse than our methods in all other metrics. LMH~\cite{2019don}, LMH-MFE~\cite{2020mfe} and LMH-CSS~\cite{2020counterfactual} surpass other methods in Num., and LMH-CSS even slightly outperforms GGE-DQ in overall Accuracy due to high performance in Num. (40.73\%). Comparing LM and LMH, it is obvious that the performance gains in Num. are due to the additional regularization for entropy. However, methods with entropy regularization drop nearly 10\% on VQA v2. This indicates that these models may over-correct the bias and largely use ``inverse language bias".

\begin{figure}[t]
	\begin{center}	
		\includegraphics[width=0.95\linewidth]{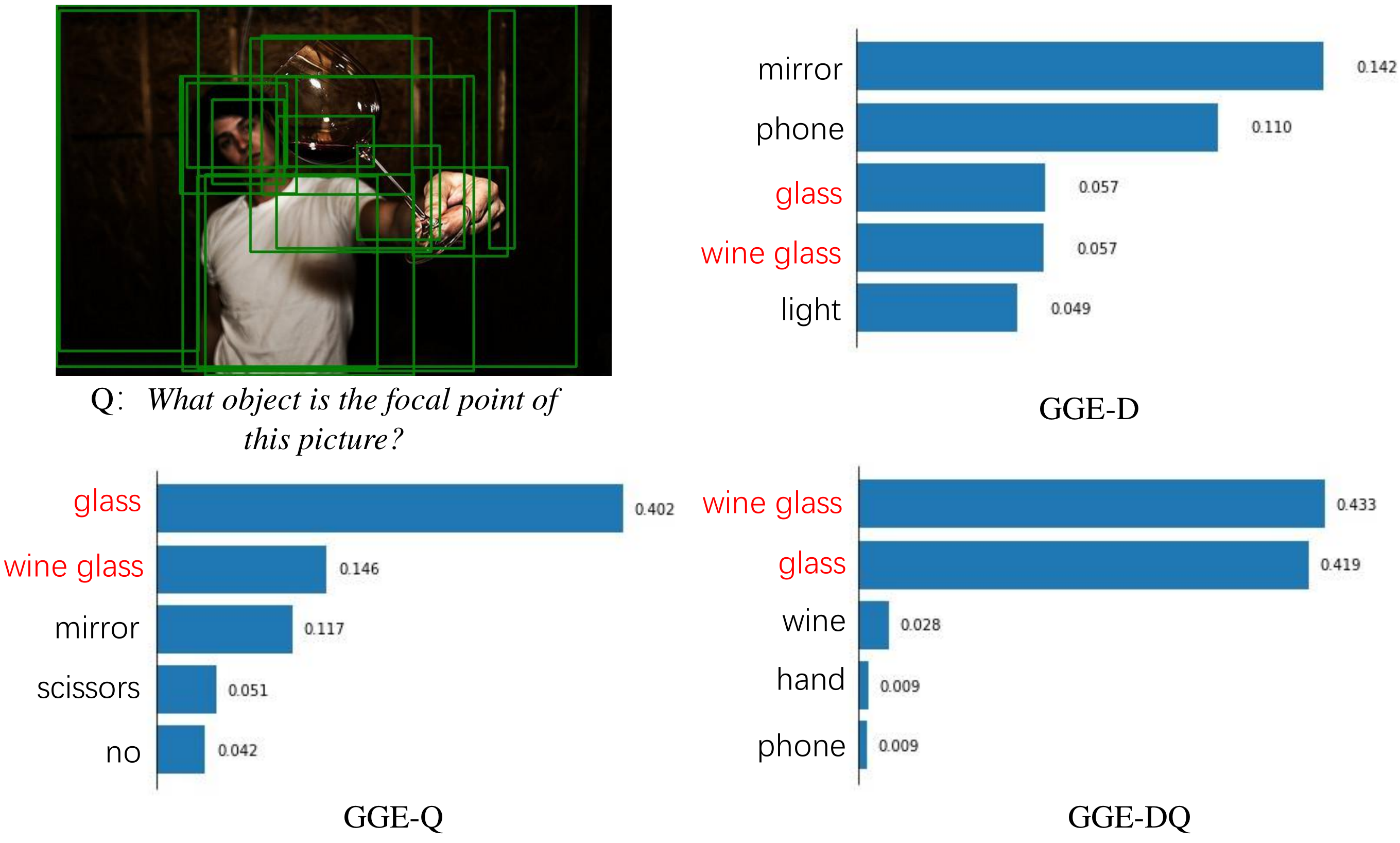}
		\vspace{-1em}
	\end{center}
	\caption{Predicted distribution for three variants of GGE.}
	\label{fig:ab}
	\vspace{-1.em}
\end{figure}

\subsection{Ablation Studies}
In this section, we design various ablations to verify the effectiveness of greedy ensemble and our claim that distribution bias and question shortcut bias are two aspects of language bias.
More results on VQA v2 are provided in Section~D in the Supplementary.

The first group of ablations is to verify if the greedy ensemble can guarantee biased data is learned with biased models. We compare with other two ensemble strategies.
{\bf SUM-DQ} directly sums up the outputs of biased models and the base model. 
{\bf LMH+RUBi} combines LMH~\cite{2019don} and RUBi~\cite{2019rubi}. It reduces distribution bias with LMH and shortcut bias with RUBi. The implementation details for these two ablations are provided in Section~C in Supplementary.

As shown in Table~\ref{tab:ablation}, SUM-DQ performs even worse than baseline. Meanwhile, the Accuracy of LMH+RUBi is just similar to that of LMH, and about $6\%$ worse than GGE-DQ.
This shows that GGE can really force the biased data to be sequentially learned with biased models. Instances that are easy to predict based on distribution or shortcut bias will be well fitted by the corresponding biased model. As a result, the base model has to pay more attention to hard examples and consider more visual information for final decision. 

In the second group of experiments, we experimentally compare distribution bias and shortcut bias. The case analysis in Figure~\ref{fig:ab} shows that GGE-D only uniforms predictions, which mainly improves Y/N as shown in Table~\ref{tab:ablation}. $B_q$ works like ``hard example mining" but will also introduce some noise (\eg ``mirror" and ``no" in this example) due to inverse distribution bias. Reducing $B_d$ at the first stage can further encourage the discovery of the hard examples and force the base model to capture visual information. In Figure~\ref{fig:ab}, the correct answer has higher confidence and the top predictions are all based on the image.
As shown in Table~\ref{tab:ablation}, GGE-DQ surpasses single-bias versions by $\sim$10\%. This well verifies our claim that distribution bias and shortcut bias are two complementary aspects of language bias.  

\begin{table}[t]
	\caption{Ablation study for different versions of GGE on VQA-CP v2 test set. \textbf{Best} results are highlighted in the columns.}
	\label{tab:ablation}
	\begin{tabular}{l|lcccc}
		\hline
		Method      & All & Y/N & Others & Num. & CGD \\ \hline
		Baseline & 39.89 & 43.01 & 45.80 & 11.88 & 3.91    \\
		SUM-DQ      & 35.46 & 42.66 & 38.01 & 12.38 & 3.10    \\
		LMH+RUBi    & 51.54 & 74.55 & 47.41 & 22.65 & 6.12    \\ \hline
		GGE-D       & 48.27 & 70.75 & 47.53 & 13.42 & 14.31    \\
		GGE-Q-iter  & 43.72 & 48.17 & 48.78 & 14.24 & 6.70    \\
		GGE-Q-tog   & 44.62 & 47.64 & 48.89 & 14.34 & 6.63  \\ \hline
		GGE-DQ-iter & 57.12 & {\bf 87.35} & {\bf 49.77} & 26.16 & {\bf 16.44}    \\
		GGE-DQ-tog  & {\bf57.32} & 87.04 & 49.59 & {\bf 27.75} & 15.27    \\ \hline
	\end{tabular}
	\vspace{-1em}
\end{table}

\begin{figure*}[t]
	\begin{center}
		\includegraphics[width=0.99\linewidth]{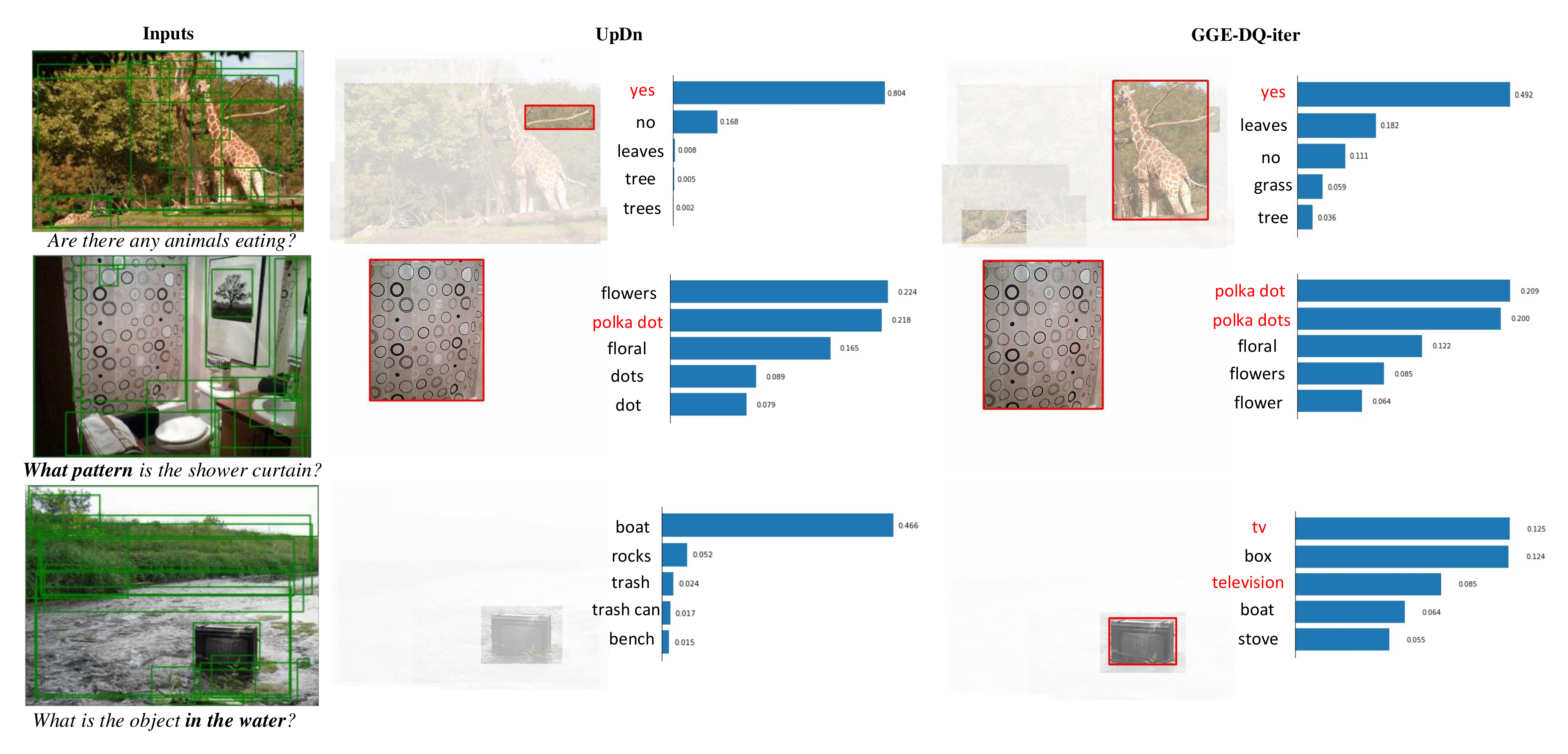}
	\end{center}
	\vspace{-1.em}
	\caption{{\bf Qualitative Evaluation for GGE-DQ}. We provide a comparison between UpDn and GGE-DQ on the visualization of the most sensitive regions and confidence of the top-5 answers. Red answers denote the ground-truth.}
	%The upper row is an example of shortcut bias, the middle row is for distribution bias and the bottom row is a more specific example for language bias. Additional results can be found in Supplementary Material.}
	\label{fig:vis}
	\vspace{-1.em}
\end{figure*}

\subsection{Generalization of GGE}
{\bf Self-Ensemble}. The performance of GGE largely depends on the predefined biased features, which requires prior knowledge of the task or dataset.% (\eg distribution bias and shortcut bias for VQA-CP).
In order to further discuss the generalization of GGE, we test a more flexible Self-Ensemble fashion (GGE-SF) on VQA-CP. GGE-SF takes the joint representation $r_i = m\left(e_v(v_i), e_q(q_i) \right)$ itself as the biased feature instead of predefined question-only branch, the biased prediction is 
\begin{equation}\label{sf}
B_{s_i} = c_s \left(r_i  \right),
\end{equation}
where $c_s: r \rightarrow \mathbb{R}^C$ is the classifier of the biased model. The training process is the same as GGE-Q.

As shown in Table~\ref{tab:generalization}, GGE-SF still surpasses the baseline even without predefined biased features. This means that the base model itself can also be regarded as a biased model, as long as the tasks or datasets are biased enough. Moreover, if we first remove distribution bias with GGE-D before Self-Ensemble, the performance of GGE-D-SF is also comparable to existing state-of-the-art methods.
\begin{table}[t]
	\caption{Variants of GGE on VQA-CP v2. SF stands for Self-Ensemble, $sxce$ denotes models trained with softmax+CE loss.}
	\label{tab:generalization}
	\setlength{\tabcolsep}{2.5mm}
	\begin{tabular}{l|lcccc}
		\hline
		Method      & All & Y/N & Others & Num.  \\ \hline
		UpDn           & 39.89 & 43.01 & 45.80 & 11.88    \\
		UpDn$_{sxce}$  & 41.37 & 45.96 & 46.90 & 12.46    \\ 
		\hline
		GGE-SF-iter    & 44.53 & 50.98 & 48.90 & 18.24    \\
		GGE-SF-tog     & 43.10 & 49.90 & 47.33 & 17.74  \\
		GGE-D-SF-iter  & 56.33 & 86.43 & 49.32 & 24.37 \\
		GGE-D-SF-tog   & 52.86 & 76.25 & 49.46 & 20.56  \\
		\hline
		GGE$_{sxce}$-D       & 53.98 & 86.06 & 47.85 & 15.09 \\
		GGE$_{sxce}$-Q-iter  & 52.98 & 82.27 & 48.06 & 14.97    \\
		GGE$_{sxce}$-Q-tog   & 52.99 & 81.86 & 47.97 & 16.11  \\
		GGE$_{sxce}$-DQ-iter   & 56.25 & 85.08 & 48.56 & 24.78 \\
		GGE$_{sxce}$-DQ-tog    & 55.84 & 84.47 & 48.76 & 26.96 \\
		\hline
	\end{tabular}
	\vspace{-1em}
\end{table}

{\bf Generalization for Loss Function}. For a fair comparison with previous work, we adopt Sigmoid+BCE loss for the above experiments. Actually, GGE is agnostic for classification losses. We provide extra experiments for Softmax+CE loss in Table~\ref{tab:generalization}. 
The implementation for GGE$_{sxce}$ is provided in Section~A in the Supplementary.

{\bf Generalization for Base Model}. GGE is also agnostic for base model choices. We provide extra experiments with BAN~\cite{2018BAN} and S-MRL~\cite{2019rubi} as base model. The results are provided in Section~D in the Supplementary.

\subsection{Qualitative Evaluation}
Examples in Figure~\ref{fig:vis} illustrate how GGE-DQ makes of visual information for inference. From top to bottom, we provide three representative failure cases from baseline UpDn.
The first example is about shortcut bias. Despite offering the right answer ``yes", the prediction from UpDn is not based on the right visual grounding. On the contrary, GGE correctly grounds the giraffe that is eating leaves. The second example is about distribution bias. UpDn correctly grounds the curtain but still answers the question based on distribution bias (``flowers" is the most common answer for ``what pattern..." in the train set). The last example is a case for reducing language prior apart from Yes/No questions. UpDn answers ``boat" just based on the language context ``in the water", while GGE-DQ provides correct answers ``tv" and ``television" with more salient visual grounding. These examples qualitatively verify our improvement in both Accuracy and visual explanation for the predictions. More examples and failure cases can be found in Supplementary.

%\subsection{Discussion}
%There are still some issues about language bias that deserves further consideration. First, we find that GGE also suffers from degradation on in-distribution data (VQA v2) similar to previous ensemble-based methods. This indicates that the model may over-estimate the bias for some instances. Similarly, according to Figure~\ref{fig:vis}, the confidence of our predictions cannot recover the answer distribution very well. We speculate that it is due to the small scale of gradient for some samples. %As shown in Eq.~\ref{direction}, the pseudo label is always less than 1, which means GGE only weakens samples that are easy for biased models but does not obviously strengthen hard samples. 
%How to control the over-fitting ``degree" of based models and scale up pseudo labels are potential research directions in the future. 

\section{Conclusion}
In this paper, we experimentally analyse several methods for robust VQA and propose a new framework to reduce the language bias in VQA. We demonstrate that the language bias in VQA can be decomposed into distribution bias and shortcut bias and then propose a Greedy Gradient Ensemble strategy to removes such two kinds of preferences step by step. % The idea of GGE is to take advantage of the over-fitting in deep learning for bias inhibition.
Experimental results demonstrate the rationality of our bias decomposition and the effectiveness of GGE. 
We believe the idea behind GGE is valuable and has the potential to become a generic method for dataset bias problems. In the future, we will 
extend GGE to solve bias problems for other tasks, provide a more rigorous analysis to guarantee model convergence, and learn to automatically detect different kinds of bias features without prior knowledge.

\section*{Acknowledgement} 
%\noindent {\bf Acknowledgement}
This work was supported in part by the National Key R\&D Program of China under Grant 2018AAA0102003, in part by National Natural Science Foundation of China: 62022083, 61620106009, 61836002 and 61931008, in part by Key Research Program of Frontier Sciences, CAS: QYZDJ-SSW-SYS013, and in part by the Beijing Nova Program under Grant Z201100006820023. Authors are grateful to Kingsoft Cloud for free GPU computing support.

%%%%%%%%%%%------------------------------------------------------------------------%

\clearpage

{\small
\bibliographystyle{ieee_fullname}
\bibliography{egbib}
}

\clearpage

{\section*{\Large Appendix}}
\setcounter{equation}{0}
\setcounter{subsection}{0}
\setcounter{section}{0}
\renewcommand\thesection{\Alph{section}}

%This supplementary document is organized as follows:
%
- Section A introduces A.1 GGE for Sigmoid+BCE loss (Section~4.1); A.2 GGE for Softmax+CE loss (Section~5.4); A.3 algorithm for GGE-iter and GGE-tog (Section~4.1).

- Section B provides more detailed settings for CGR, CGW, and CGD (Section 5.1).

- Section C provides C.1 implementation details for the base model; C.2 and ablations for ensemble strategy, SUM-DQ and LMH+RUBi (Section~5.3).

- Section D provides D.1 ablation studies for base model S-MRL and BAN (Section~5.3); D.2 comparison between Self-Ensemble GGE and RUBi; D.3 additional experimental results (Section~3.2 and 5.1), including Accuracy on VQA v2 and CGR/CGD for all implemented methods.

- Section E provides more quantitative examples and failure cases from GGE-DQ (Section 5.4).

%\begin{itemize}
%	\item Section A introduces a more detailed analysis of the gradient used in GGE (Section 4).
%	\item Section B provides additional experimental results (Section 3 and 5).
%	\item Section C provides more quantitative examples and failure cases from GGE-DQ (Section 5).
%	\item Codes are also provided in Code.zip.
%\end{itemize}

\section{Implementation Details for GGE}
\subsection{Sigmoid+BCE}
For classification problem with BCE loss, the negative gradient is shown in Eq.~7 in the main paper
\begin{equation}\label{direction}
	-\nabla \mathcal{L}(\mathcal{H}_{m,i}) %:= \frac{\partial \mathcal{L}\left(\mathcal{H}_m, Y  \right)}{\partial \mathcal{H}_{m,i}} 
	= 2y_{m,i} \sigma \left(-2 y_{m,i} \mathcal{H}_{m,i} \right).
\end{equation}

If $y_{m,i}=0$, the gradient will always be 0.
If the label $y_{m,i}=1$, we plot the change of negative gradient versus prediction $\mathcal{H}_{m,i}$. 

\begin{figure}[t]
	\begin{center}
		\subfigure[$-\nabla \mathcal{L}(\mathcal{H}_{m,i})$ vs. $\mathcal{H}_{m,i}$]{
			\includegraphics[width=0.9\linewidth]{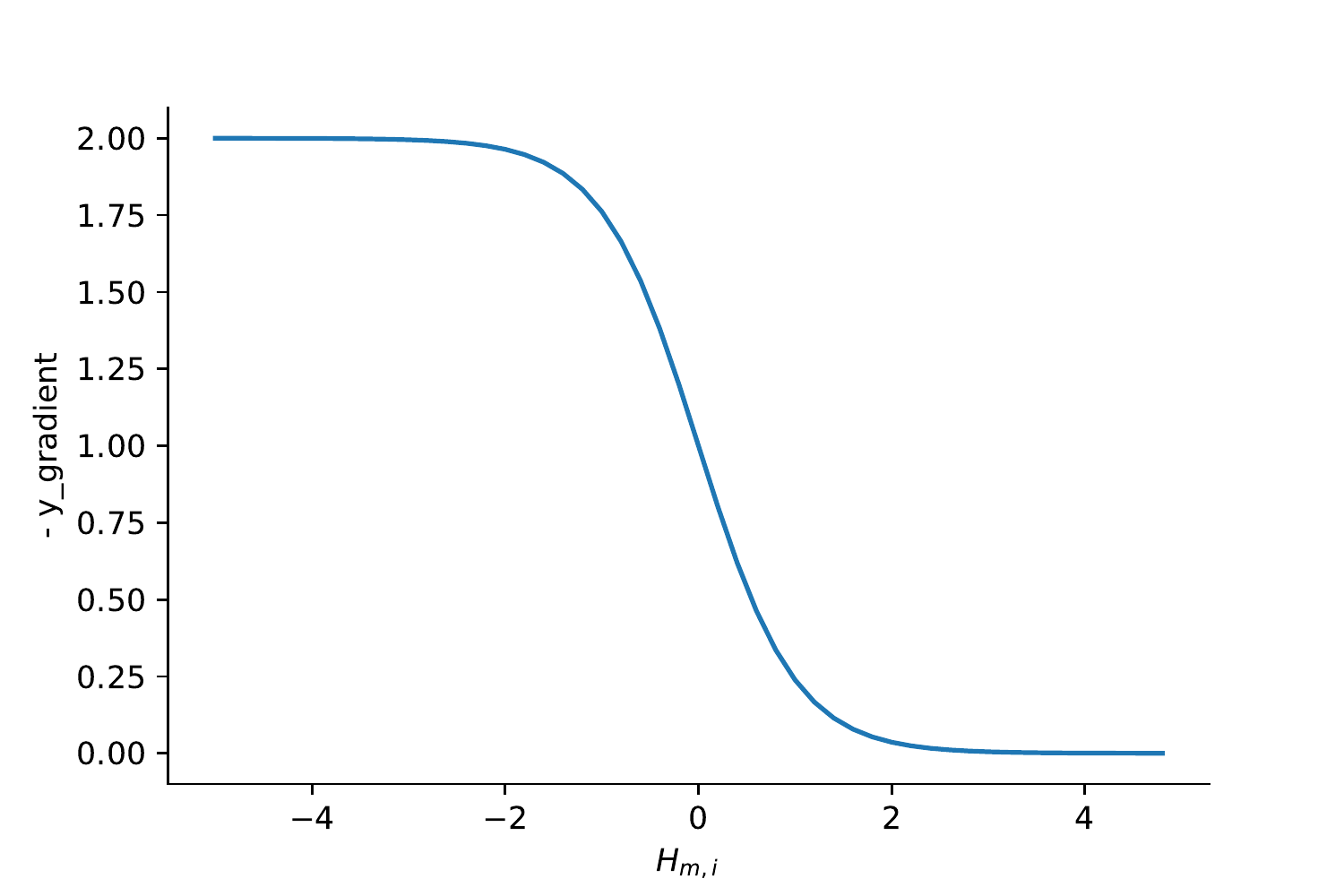}
		}\\ 
		\vspace{-1em}
		\subfigure[$-\nabla \mathcal{L}(\mathcal{H}_{m,i})$ vs. $\sigma(\mathcal{H}_{m,i})$]{
			\includegraphics[width=0.9\linewidth]{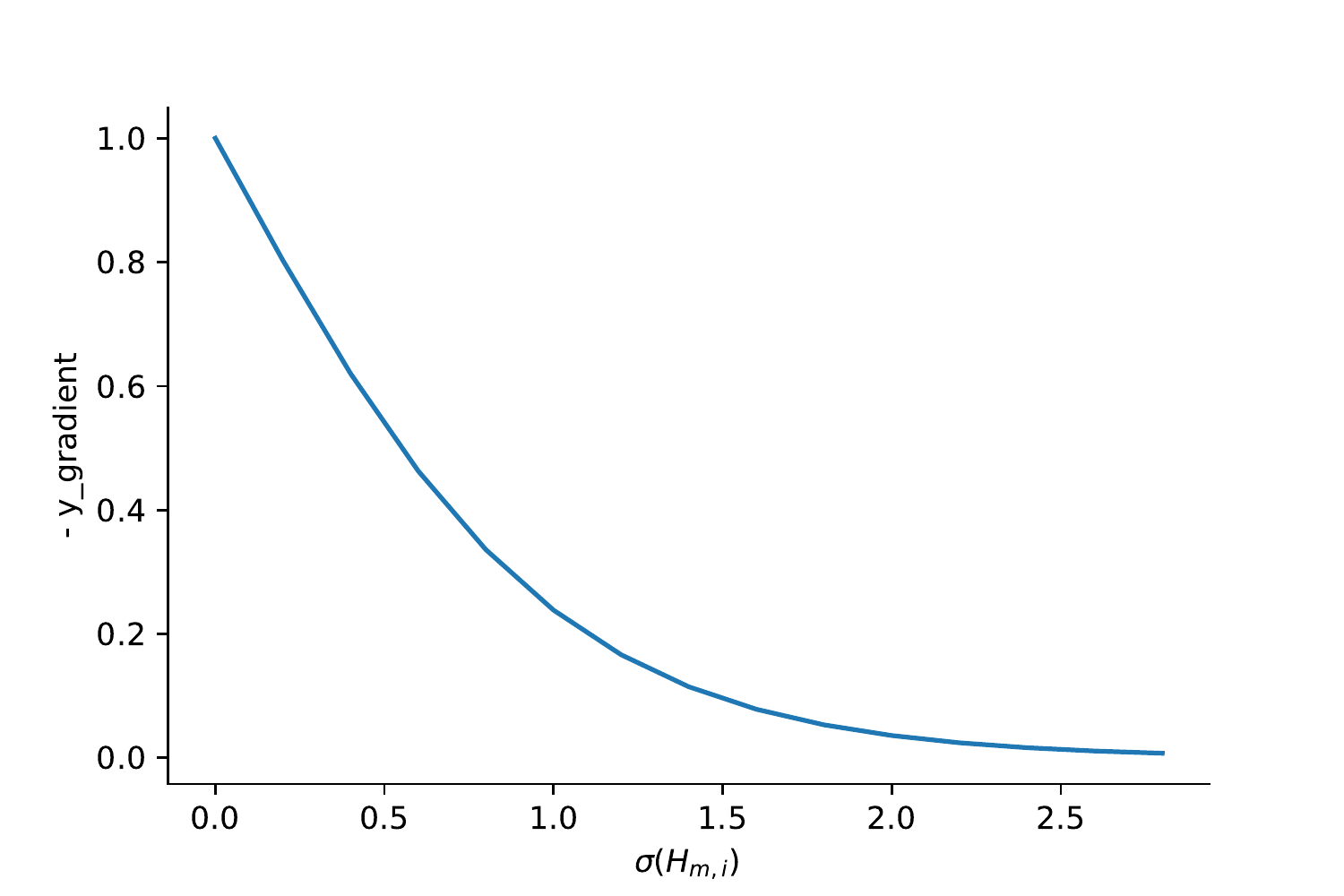}
		}
	\end{center}
	\caption{{\bf Negative Gradient versus Predictions}}
	\label{plot}
	\vspace{-2em}
\end{figure}

As shown in Figure~\ref{plot}, the gradient will continuously decrease when biased models can predict the right answers with higher confidence. This means the base model will pay more attention to samples that are hard to solve by biased models.

In practice, we clip $\sigma(\mathcal{H}_{m,i}) > 0$ with Sigmoid function, to make the range of $-\nabla \mathcal{L}(\mathcal{H}_{m,i})$ consistent with the label space [0,1] of BCE loss. $B_d$ is a statistic answer distribution of the training set, which satisfies $B_d >0 $. Therefore, we do not need to add Sigmoid function on distribution bias in Eq.11-15 in the main paper. 

However, clipping the gradient does not directly increase the scale of hard samples but only lowers the scale of easy ones, resulting in performance degradation on VQA v2. Actually, for the hard samples, the gradient can be up to 2.0 without clip operation. If we can design a new classification loss with label space $[0, 2]$ in place of BCE, it may be an alternative approach to deal with this problem.

\subsection{Softmax+CE}
We provide GGE optimized with Softmax+CE loss in Section~5.4. The loss function can be written as
\begin{equation}\label{ce}
	\mathcal{L}(Z, Y) = -\sum_{i=1}^{C} y_i \log(\sigma_i),
\end{equation}
with
\begin{equation}\label{softmax}
	\sigma_i = \frac{e^{z_i}}{\sum_{j=1}^{C}z_j},
\end{equation}
where $Z = \{z_i\}_{i=1}^{C} $ is the predicted logits, $C$ is the number of classes, and $y_i \in [0,1]$ is the ground truth labels.
The negative gradient of loss function is 
\begin{equation}\label{g_sfce}
	-\nabla \mathcal{L}(z_i) = y_i - \sigma_i.
\end{equation}
Similar to implementation of Sigmoid+BCE, we directly clip the $\nabla L(z_i)$ to the label space [0,1]
\begin{equation}\label{se}
	-\nabla \mathcal{\hat{L}}(z_i) = 
	\begin{cases}
		y_i - \sigma_i & y_i > 0\\
		0              & y_i = 0
	\end{cases}.
\end{equation}
As a result, if $y_i = 0$ the pseudo label $\mathcal{\hat{L}}(z_i)$ will still be 0, otherwise, it will decrease when biased models can predict the right answer with higher confidence. The optimization process is the same with that in Sigmoid+BCE. Additionally, since the statistical distribution $B_d \in (0,1)$, we treat $\sigma_i = B_{d_i}$ when calculate the gradient in GGE-D and GGE-DQ. 

\subsection{GGE-Iter and GGE-tog}
In Section~4.1 we provide two optimization schemes GGE-iteration and GGE-together. The detailed implementation is shown in Algorithm~\ref{alg:iter} and~\ref{alg:tog}. Two variants of implementation do not show an obvious performance gap in most experiments.

\begin{algorithm}[t]
	\caption{GGE-iteration}
	\label{alg:iter}
	%	\LinesNumbered 
	\KwIn{ Observations $X$, Lables $Y$,\\
		Biased features Observations $\mathcal{B} = \{B_m\}_{m=1}^{M} $, \\
		Base function $f(.|\theta): X \rightarrow \mathbb{R}^{|Y|}$,\\
		Bias functions $\{h_m(.|{\phi_i}): B_i \rightarrow \mathbb{R}^{|Y|} \}_{m=1}^{M} $ }
	{\bf Initialize: $\mathcal{H}_0 = 0 $ }  \;
	\For{Batch $t= 1 \dots T$}{
		\For{$m = 1 \dots M$}{
			%$\hat{Y} \leftarrow -\nabla \mathcal{L}(H_{m-1}, Y) $ \\
			$L_m(\phi_m) \leftarrow  \mathcal{L}' \left(h_{m}(B_{m};\phi_{m}),-\nabla \mathcal{L}(H_{m-1}, Y) \right) $\\
			Update $\phi_m \leftarrow \phi_m - \alpha \nabla_{\phi_m} L_m(\phi_m) $\\
		}
		%$\hat{Y} \leftarrow -\nabla \mathcal{L}(H_M, Y) $ \\
		$L_{M+1}(\theta) \leftarrow  \mathcal{L}' \left(f(X;\theta), -\nabla \mathcal{L}(H_M, Y) \right) $ \\
		Update $\theta \leftarrow \theta - \alpha \nabla_{\theta} L_{M+1}(\theta) $\\
	} 
	\Return $ Y = f(X; \theta) $  
	%	$\tilde{a} \leftarrow \text{MLP}(b_T, h_e^T) $\\
	%	$\mathcal{L}(\Theta) \leftarrow - \text{BCE}(a, \tilde{a}) - \sum_{t=1}^{T}\mathcal{L}_t $\\
	%	Update model parameters $\Theta \leftarrow \Theta - \alpha \nabla_\Theta \mathcal{L}(\Theta) $
\end{algorithm}

\begin{algorithm}[t]
	\caption{GGE-together}
	\label{alg:tog}
	%	\LinesNumbered 
	\KwIn{ Observations $X$, Lables $Y$,\\
		Biased features Observations $\mathcal{B} = \{B_m\}_{m=1}^{M} $, \\
		Base function $f(.|\theta): X \rightarrow \mathbb{R}^{|Y|}$,\\
		Bias functions $\{h_m(.|{\phi_i}): B_i \rightarrow \mathbb{R}^{|Y|} \}_{m=1}^{M} $ }
	{\bf Initialize: $\mathcal{H}_0 = 0 $ }  \;
	\For{Batch $t= 1 \dots T$}{
		\For{$m = 1 \dots M$}{
			%$\hat{Y} \leftarrow  $ \\
			$L_m(\phi_m) \leftarrow  \mathcal{L}' \left(h_{m}(B_{m};\phi_{m}),-\nabla \mathcal{L}(H_{m-1}, Y) \right) $\\
		}
		%$\hat{Y} \leftarrow -\nabla \mathcal{L}(H_M, Y) $ \\
		$L_{M+1}(\theta) \leftarrow  \mathcal{L}' \left(f(X;\theta), -\nabla \mathcal{L}(H_M, Y)\right) $ \\
		$L(\Theta) \leftarrow \sum_{m=1}^{M+1} L_m $ \\
		Update $\Theta \leftarrow \Theta - \alpha \nabla_\Theta L(\Theta)$ \\
	} 
	\Return $ Y = f(X; \theta) $  
	%	$\tilde{a} \leftarrow \text{MLP}(b_T, h_e^T) $\\
	%	$\mathcal{L}(\Theta) \leftarrow - \text{BCE}(a, \tilde{a}) - \sum_{t=1}^{T}\mathcal{L}_t $\\
	%	Update model parameters $\Theta \leftarrow \Theta - \alpha \nabla_\Theta \mathcal{L}(\Theta) $
	
\end{algorithm}

\section{Details for CGD}
\begin{figure*}[t]
	\begin{center}
		\subfigure[CGR]{
			\label{subfig:CGR}
			\includegraphics[width=0.31\linewidth]{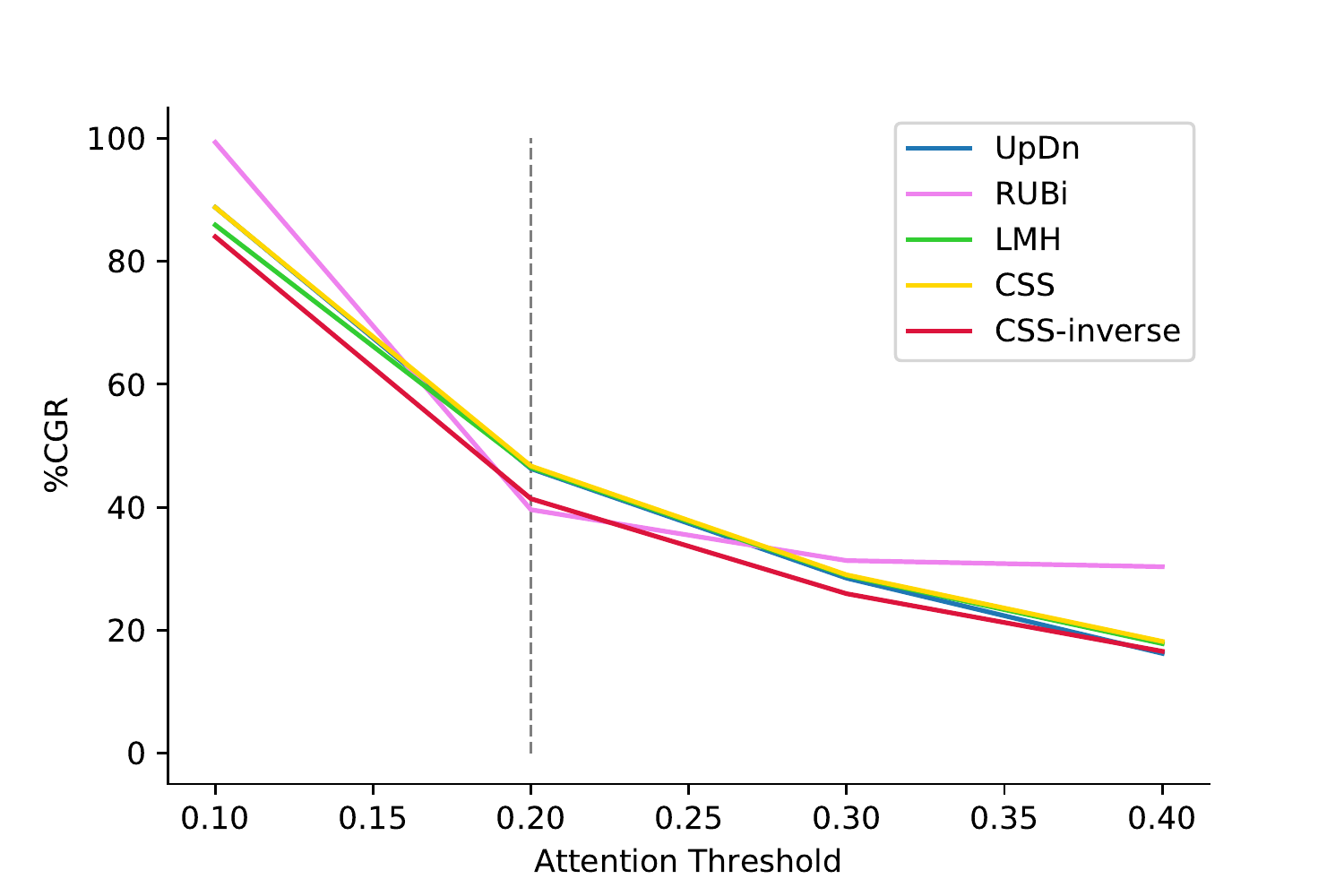}
		}
		\subfigure[CGW]{
			\label{subfig:CGW}
			\includegraphics[width=0.31\linewidth]{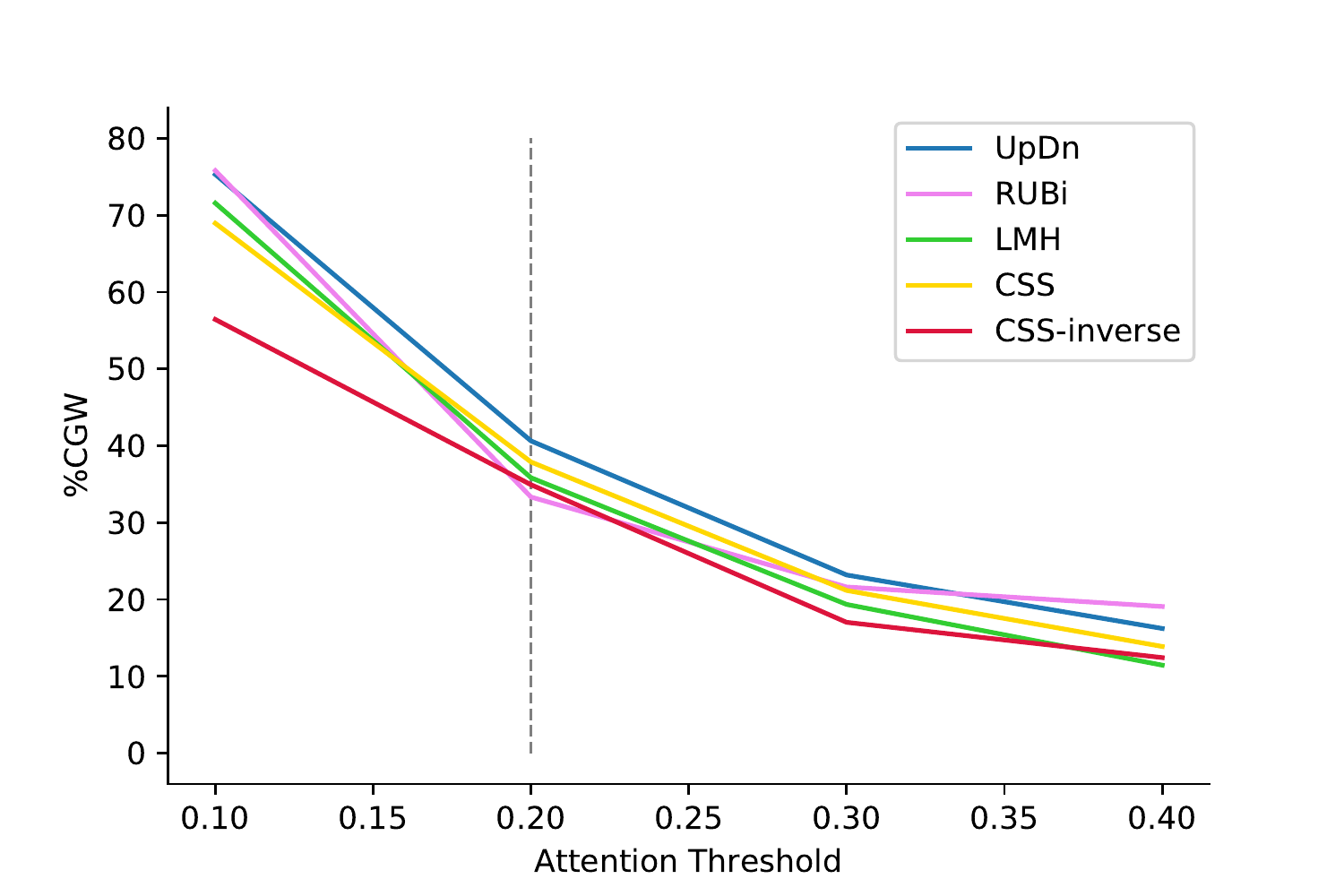}
		}
		\subfigure[CGD]{
			\label{subfig:CGD}
			\includegraphics[width=0.31\linewidth]{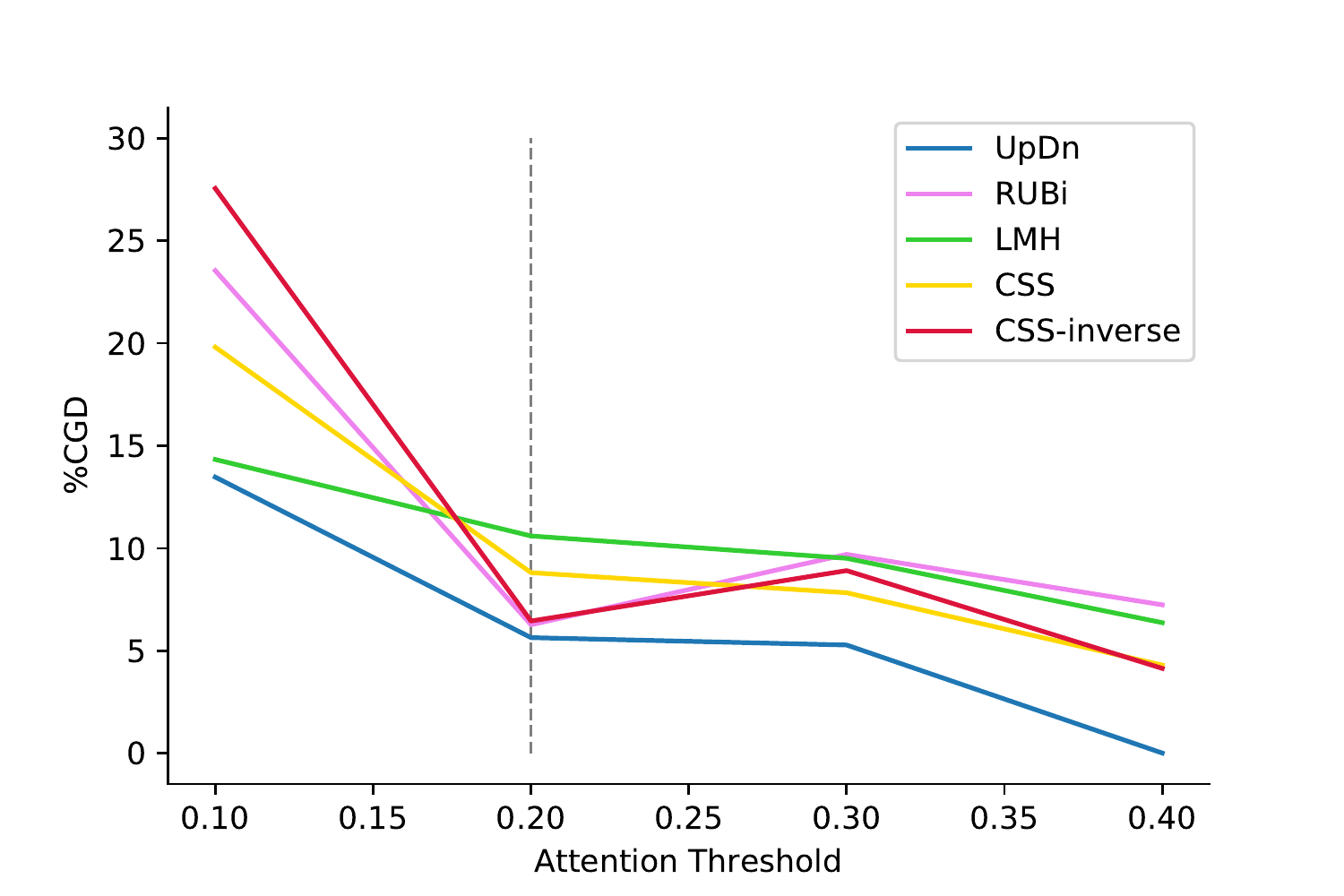}
		}
		
	\end{center}
	\vspace{-1em}
	\caption{{\bf CGR, CGW and CGD versus attention threshold for prevailing methods}. }
	\label{fig:CGD}
	\vspace{-1.5em}
\end{figure*}

First, we should stress that CGD \emph{only} evaluates whether the visual information is taken for answer prediction, which is \emph{parallel} with Accuracy and different from metrics in Referring Expression and Visual Grounding tasks. It is proposed to help quantitatively evaluate models' grounding ability.

CGD considers the top-\emph{N} most sensitive visual region. In this paper, we evaluate the sensitivity via attention. In Figure~\ref{fig:CGD}, we plot change of CGR, CGW and CGD with different threshold for prevailing methods UpDn~\cite{2018bottomup}, RUBi~\cite{2019rubi}, LMH~\cite{2019don} CSS~\cite{2020counterfactual} and CSS-V$_{inv-hat}$. We set attention threshold $t \in \{ 0.1, 0.2, 0.3, 0.4\}$, which indicates that top-\emph{N} is no more than \{9, 4, 3, 2\}.  

We choose to consider top-4 ($t=0.2$) objects for CGD, since many questions need to consider multiple objects and $t=0.2$ is the most discriminative threshold as shown in Figure~\ref{subfig:CGD}. Apart from attention, Grad-CAM~\cite{2017gradcam} can be an alternative for grounding evaluation.

\section{Implementation Details for Experiments}
\subsection{Base Model}
We use the publicly available reimplementation of UpDn\footnote{https://github.com/hengyuan-hu/bottom-up-attention-vqa}~\cite{2018bottomup} for our baseline architecture, data preprocess and optimization.

{\bf Image Encoder}. Following the popular bottom-up attention mechanism~\cite{2018bottomup}, we use a Faster R-CNN~\cite{2015faster-rcnn} based framework to extract visual features. We select the top-36 region proposals for each image $\mathbf{v} \in \mathbb{R}^{36\times 2048}$.

{\bf Question Encoder}. Each word is first initialized by 300-dim GloVe word embeddings~\cite{2014glove}, then fed into a GRU with 1024-d hidden vector. The question representation is the last state of GRU $h_T \in \mathbb{R}^{1024}$.

{\bf Multi-modal Fusion}. We use traditional linear attention between $h_T$ and $\mathbf{v}$ for visual representation. and the final representation for classification is the Hadamard product of vision and question representation.

{\bf Question-only Classifier}. The question-only classifier is implemented as two fully-connected layers with ReLU activations. The input question representation is shared with that in VQA base model.

{\bf Question types}. We use 65 question types annotated in VQA v2 and VQA-CP, according to the first few words of the question (e.g., ``What color is"). To save the training time, we simply use statistic answer distribution conditioned by question type in the train set as the prediction of distribution bias.

{\bf Optimization}. Following UpDn~\cite{2018bottomup}, all the experiments are conducted with the Adamax optimizer for 20 epochs with learning rate initialized as 0.001. We train all models on a single RTX 3090 GUP with PyTorch 1.7~\cite{2019pytorch} and batch size 512.

{\bf Data Preprocessing}. Following previous works, we filter the answers that appear less than 9 times in the train set. For each instance with 10 annotated answers, we set the scores for labels that appear 1/2/3 times as 0.3/0.6/0.9, more than 3 times as 1.0.

\subsection{Ablations for Ensemble}
{\bf SUM-DQ}. SUM-DQ ablation is to verify if GGE can learn biased data with biased models.
The loss for the whole model is
\begin{equation}\label{sum_dq}
	L = \mathcal{L}(B_d + \sigma(B_q) + \sigma(\tilde{A}), A).
\end{equation}

{\bf LMH+RUBi}. LMH~\cite{2019don} and RUBi~\cite{2019rubi} are methods that can only reduce a single type of bias. LMH+RUBi is a direct combination of LMH and RUBi. It reduces distribution bias with LMH and shortcut bias with RUBi step by step.
The loss for RUBi is written as
\begin{equation}\label{rubi}
	L_{rubi}(\tilde{A}, A) = \mathcal{L}(\tilde{A} \odot \sigma(G_q), \tilde{A}) + \mathcal{L}(c_q(G_q), A ),
\end{equation}
where $G_q = g(e_q(q_i))$, $g(.)Q \rightarrow \mathbb{R}^C$. Combining with LMH, we compose $A$ as
\begin{equation}\label{lhm}
	F(A,B,M) = \log A + g(M)\log B,
\end{equation}
where $M$ and $B$ are the fused feature and the bias in LMH. The combined loss function is
\begin{equation}\label{lhm_rubi}
	L = L_{rubi}(F(A,B,M), \tilde{A}) + wH(g(M)\log B),
\end{equation} 
where $H(.)$ is the entropy and $w$ is a hyper-parameter.

\section{Supplementary Experimental Results}

\subsection{Ablations of Base Models}
We do experiments on other base models BAN~\cite{2018BAN} and S-MRL~\cite{2019rubi}. The models are re-implemented based on officially released codes. For BAN, we set the number of Bilinear Attention blocks as 3. We choose the last bi-linear attention map of BAN and sum up along the question axis, which is referred to as the object attention for CGR and CGW. Although Accuracy of our reproduced S-MRL is a litter lower than that in \cite{2019rubi}, GGE-DQ can improve the Accuracy over 10\% and surpass most of the existing methods. As shown in the table, GGE is a model-agnostic de-bias method, which can improve all three base models UpDn~\cite{2018bottomup}, S-MRL\cite{2019rubi} and BAN~\cite{2018BAN} by a large margin.

\subsection{Self-Ensemble Comparison}
We provide an additional experiment for RUBi~\cite{2019rubi} with Self-Ensemble fashion. The input of the question-only branch is replaced by the joint representation from the base model. As shown in Table~\ref{tab:my-table}, RUBi-SF is even worse than baseline UpDn on both VQA-CP v2 and VQA v2. On the contrary, Accuracy of GGE-SF is comparable to GGE-Q, which further demonstrates the generalization of GGE.

\subsection{Additional Experimental Results}
We provide detailed CGR, CGW, and results on VQA-CP and VQA v2 for all re-implemented methods in Section 3 and Section 5. 

As shown Table~\ref{tab:my-table}, GGE-DQ largely improves more challenging ``Others" question type~\cite{teney2020value}. This means that GGE-DQ really focuses on images largely rather than only relying on ``inverse language bias" for higher Accuracy.
Moreover, Inverse-Supervision strategy does not improve GGE-DQ-tog (GGE-DQ-tog$_{is}$ in Table~\ref{tab:my-table}), which also demonstrates that GGD-DQ better reduces distribution bias compared with other methods.

There are still some issues about language bias that deserves further consideration. 
First, both GGE-D$_{sxce}$ and GGE-Q$_{sxce}$ are robust on VQA v2 but GGE-DQ$_{sxce}$ drops a lot. We think the softmax function will amplify the gradient of biased models and over-estimate the dataset biases.
Second, LMH+RUBi performs much better than both LMH and RUBi on VQA v2. This can bring further research into the relationship between distribution bias and shortcut bias. Third, UpDn$_{is}$ does not degrade a lot in VQA v2, which indicates some entanglement between entropy regularization and Inverse-Supervision strategy. 

Moreover, we find that GGE also suffers from degradation on in-distribution data (VQA v2) similar to previous ensemble-based methods. This indicates that the model may over-estimate the bias for some instances. We speculate that it is due to too small scale of the gradient for some samples easy to fit by distribution bias or shortcut bias. 
How to control the over-fitting ``degree" and scale up pseudo labels are potential research directions in the future. 

\section{Additional Qualitative Results}
\begin{figure*}[t]
	\begin{center}
		\includegraphics[width=0.99\linewidth]{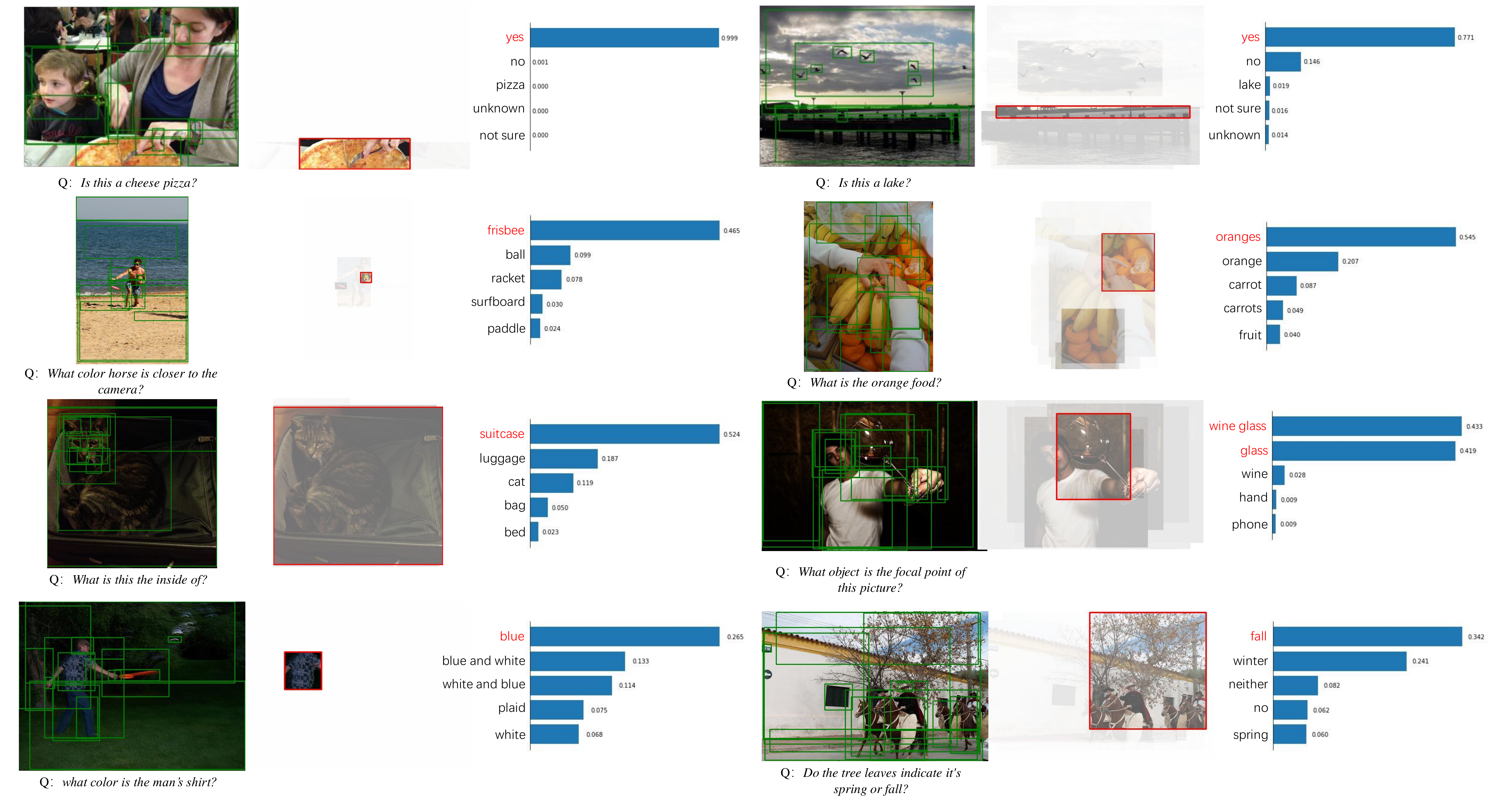}
	\end{center}
	\caption{{\bf More examples from GGE-DQ}. The model can successfully provide the right prediction with right evidences.}
	\label{visual}	
\end{figure*}

\begin{figure*}[t]
	\begin{center}
		\includegraphics[width=0.99\linewidth]{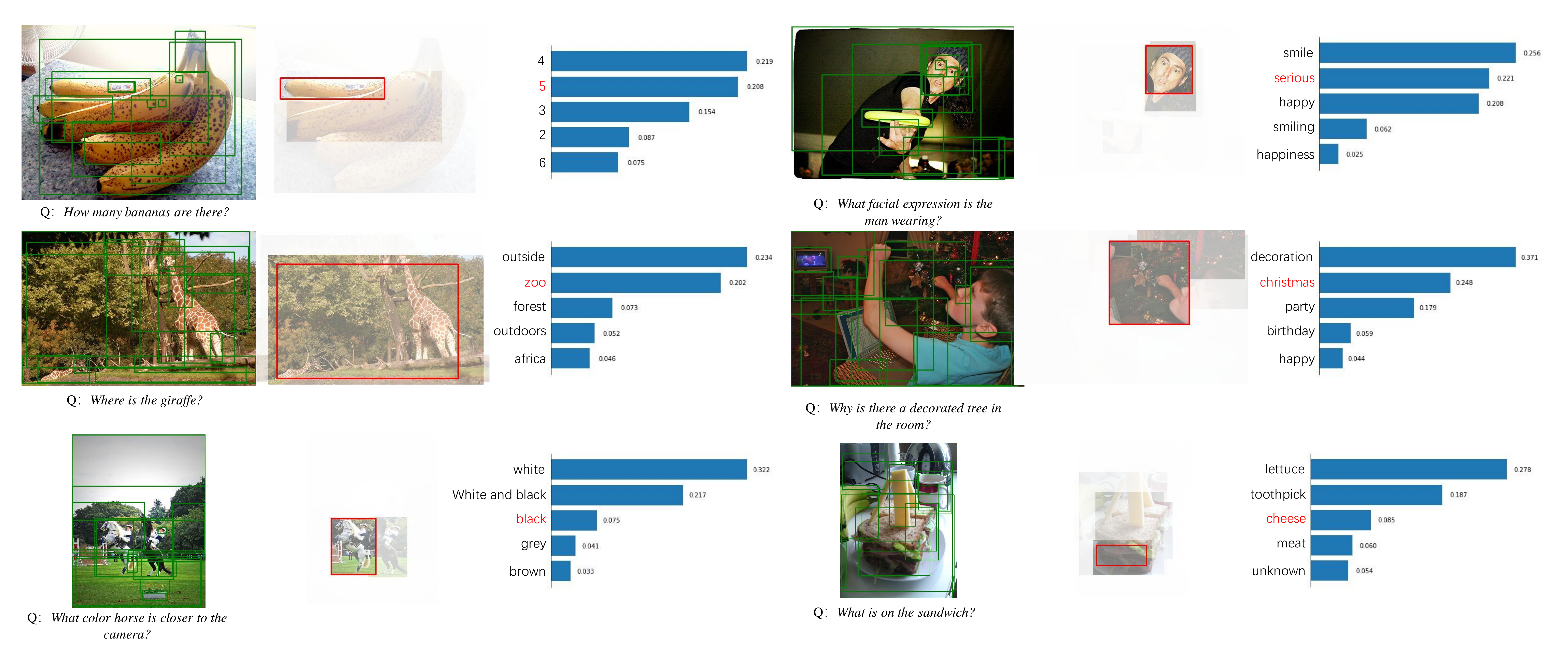}
	\end{center}
	\caption{{\bf Failure Cases}. Most of the failure cases still match their visual explanations (Wrong predictions with corresponding wrong evidences). The model is still weak in counting problem and questions that hardly appear in the train set (upper row). Some failure case are due to missing annotation in the dataset, since ``outside" and ``decoration" can also be regarded as the right answers (middle row). 
		The last row shows that answers for failure cases are still consistent with visual explanations rather than language bias, which is identified by low CGW and indicates GGE-DQ really has better visual-grounding ability.
	}
	\label{failure}
\end{figure*}

In this section, we provide more examples from GGE-DQ in Figure~\ref{visual} and some failure cases in Figure~\ref{failure}. 

\clearpage
\begin{table*}[t]
	\centering
	\setlength{\tabcolsep}{4mm}
	\caption{{\bf Ablations of base model BAN and S-MRL}.}
	\label{tab:ablation}
	\begin{tabular}{lccccllc}
		\hline
		\multirow{2}{*}{Method} & \multicolumn{7}{c}{VQA-CP test}             \\ \cline{2-8} 
		& All & Y/N & Num. & Others & $\uparrow$CGR & $\downarrow$CGW & $\uparrow$CGD \\ \hline
		S-MRL~\cite{2019rubi}   & 37.90 & 43.68 & 12.04 & 41.97 & \textbf{41.94} & 27.32 & 14.62  \\
		+GGE-DQ-tog   & \textbf{54.62} & 76.11 & 18.04 & \textbf{47.70} & 35.61 & \textbf{18.17} & \textbf{17.44} \\
		+GGE-DQ-iter  & 54.03 & \textbf{79.66} & \textbf{20.77} & 46.72 & 38.10 & 22.42 & 15.68  \\ \hline
		BAN~\cite{2018BAN}   & 35.94 & 40.39 & 12.24 & 40.51 & 5.33 & 5.19 & 0.14  \\
		+GGE-DQ-tog   & \textbf{51.91} & \textbf{81.37} & 21.85 &45.46 &  \textbf{36.93} & 27.10 & \textbf{9.83}  \\
		+GGE-DQ-iter   & 50.75 & 74.56 & 20.59 & \textbf{46.54} & 20.87 & \textbf{16.85} & 4.98   \\ \hline
	\end{tabular}
\end{table*}

\begin{table*}[t]
	\small
	\centering
	\caption{{\bf Extra experimental results for Section 3 and Section 5}.  }
	\label{tab:my-table}
	\begin{tabular}{lccccllcccccc}
		\hline
		\multirow{2}{*}{Method} & \multicolumn{7}{c}{VQA-CP test}            &  & \multicolumn{4}{c}{VQA v2 val} \\ \cline{2-8} \cline{10-13} 
		& All & Y/N & Num. & Others & CGR & CGW & CGD &  & All  & Y/N  & Num. & Others \\ \hline
		UpDn~\cite{2018bottomup}           & 39.89 & 43.01 & 12.07 & 45.82 & 44.27 & 40.63 & 3.91  & & \textbf{63.79} & 80.94 & \textbf{42.51} & \textbf{55.78} \\
		HINT~\cite{2019hint}           & 47.50 & 67.21 & 10.67 & 46.80 & 45.21 & 34.87 & 10.34 & & 63.38 & \textbf{81.18} & 42.14 & 55.66 \\
		RUBi~\cite{2019rubi}           & 45.42 & 63.03 & 11.91 & 44.33 & 39.60 & 33.33 & 6.27  & & 55.19 & 61.04 & 41.00 & 54.43 \\
		LM~\cite{2019don}             & 48.78 & 70.37 & 14.24 & 46.42 & \textbf{47.30} & 35.97 & 11.33  & & 63.26 & 81.16 & 42.22 & 55.22 \\
		LMH~\cite{2019don}            & 52.73 & 72.95 & 31.90 & 47.79 & 46.44 & 35.84 & 10.60 & & 56.35 & 65.06 & 37.63 & 54.69 \\
		CSS-V~\cite{2020counterfactual}          & 57.91 & 80.36 & 50.45 & 47.83 & 42.72 & 31.28 & 11.44 & & 53.94 & 57.48 & 55.37 & 38.39 \\
		CSS~\cite{2020counterfactual}	           & 58.11 & 83.65 & 40.73 & 48.14 & 46.70 & 37.89 & 8.81  & & 53.15 & 61.20 & 37.65 & 53.36   \\ \hline
		HINT$_{inv}$   & 47.20 & 67.23 & 13.21 & 46.15 & 42.01 & 39.11 & 2.90  & & 60.33 & 74.36 & 40.31 & 55.12 \\
		CSS-V$_{inv}$  & 58.05 & 79.84 & 52.24 & 47.23 & 41.38 & 34.93 & 6.45  & & 54.39 & 58.73 & 38.81 & 55.23 \\ 
		\hline
		UpDn$_{is}$    & 42.12 & 45.81 & 12.98 & 47.02 & 44.52 & 39.59 & 4.93  & & 62.85 & 80.34 & 42.00 & 55.08 \\
		RUBi$_{is}$    & 48.16 & 72.34 & 12.69 & 45.22 & 47.55 & 33.73 & 13.83  & & 59.10 & 76.67 & 41.09 & 50.50 \\
		LMH$_{is}$    & \textbf{58.12} & 79.73 & \textbf{53.41} & 48.01 & 39.51 & 30.82 & 8.69  & & 43.29 & 33.22 & 34.14 & 53.40 \\ 
		GGE-DQ-tog$_{is}$    & 54.64 & 85.47 & 23.43 & 47.64 & 40.47 & 25.81 & 14.66  & & 57.16 & 70.43 & 38.00 & 52.13 \\ 
		\hline
		SUM-DQ         & 35.46 & 42.66 & 12.38 & 38.01 & 41.28 & 38.18 & 3.91  & & 56.85 & 81.09 & 38.55 & 43.25 \\
		LMH+RUBi       & 51.54 & 74.55 & 22.65 & 47.41 & 46.67 & 40.55 & 6.12  & & 60.68 & 77.91 & 39.10 & 53.15 \\ 
		\hline\hline
		GGE-D          & 48.27 & 70.75 & 13.42 & 47.53 & 38.79 & \textbf{24.48} & 14.31 & & 62.79 & 79.24 & 42.31 & 55.71 \\
		GGE-Q-iter     & 43.72 & 48.17 & 14.24 & 48.78 & 43.74 & 37.04 & 6.70  & & 61.23 & 78.28 & 41.42 & 53.50 \\
		GGE-Q-tog      & 44.62 & 47.64 & 14.34 & 48.89 & 45.19 & 38.56 & 6.63  & & 62.14 & 78.64 & 40.72 & 54.21 \\ 
		\hline
		GGE-DQ-iter    & 57.12 & \textbf{87.35} & 26.16 & \textbf{49.77} & 44.35 & 27.91 & \textbf{16.44} & & 59.30 & 73.63 & 40.30 & 54.29 \\
		GGE-DQ-tog     & 57.32 & 87.04 & 27.75 & 49.59 & 42.74 & 27.47 & 15.27 & & 59.11 & 73.27 & 39.99 & 54.39 \\ 
		\hline \hline
		RUBi-SF    & 37.53 & 43.27 & 14.11 & 41.07 & 39.30 & 32.66 & 7.14 & & 55.06 & 70.85 & 30.97 & 49.44 \\
		GGE-SF-iter    & 44.53 & 50.98 & 18.24 & 48.90 & 45.07 & 38.99 & 6.08 & & 60.66 & 74.93 & 41.14 & 52.95 \\
		GGE-SF-tog     & 43.10 & 49.90 & 17.74 & 47.33 & 42.40 & 35.85 & 6.55 & & 59.00 & 73.71 & 41.14 & 52.54 \\ 
		GGE-D-SF-iter    & 56.33 & 86.43 & 23.37 & 49.32 & 43.77 & 29.30 & 14.47 & & 62.03 & 80.73 & 41.79 & 53.14 \\
		GGE-D-SF-tog     & 52.86 & 76.25 & 20.56 & 49.46 & 42.48 & 30.25 & 12.23 & & 59.00 & 73.71 & 41.14 & 52.54 \\ 
		\hline
		UpDn$_{sxce}$          & 41.37 & 45.96 & 12.46 & 46.90 & 42.81 & 40.90 & 1.91 & & 63.38 & 81.26 & 43.13 & 55.14 \\
		GGE$_{sxce}$-D         & 53.98 & 86.06 & 15.09 & 47.85 & 37.45 & 30.52 & 6.93 & & 62.34 & 79.17 & 41.50 & 55.06 \\ 
		GGE$_{sxce}$-Q-iter    & 52.98 & 82.27 & 14.97 & 48.06 & 40.64 & 31.55 & 9.09 & & 61.76 & 78.57 & 42.01 & 54.20 \\
		GGE$_{sxce}$-Q-tog     & 52.99 & 81.86 & 16.11 & 47.97 & 41.01 & 32.62 & 8.39 & & 61.38 & 77.53 & 42.30 & 54.14 \\ 
		GGE$_{sxce}$-DQ-iter   & 56.25 & 85.08 & 24.78 & 48.56 & 43.13 & 29.52 & 13.61 & & 52.38 & 54.51 & 39.93 & 54.07 \\
		GGE$_{sxce}$-DQ-tog    & 55.84 & 84.47 & 26.96 & 48.76 & 41.41 & 31.02 & 10.39 & & 52.17 & 54.17 & 40.10 & 53.85 \\ 
		\hline
	\end{tabular}
	
\end{table*}

\clearpage

%\clearpage
%{\small
%	\bibliographystyle{ieee_fullname}
%	\bibliography{egbib}
%}

\end{document}